\documentclass[10pt,twocolumn,letterpaper]{article}

\usepackage{wacv}
\usepackage{times}

\usepackage{adjustbox}
\usepackage{amsmath}
\usepackage{amssymb}
\usepackage{booktabs}
\usepackage{comment}
\usepackage{enumitem}
\usepackage{float}
\usepackage{graphicx}
\usepackage{mathtools}
\usepackage{multirow}
\usepackage{multicol}
\usepackage{pifont}
\usepackage{subfig}
\usepackage{xspace}
\usepackage{xfrac}
\usepackage[table,xcdraw]{xcolor}
\usepackage[colorlinks,pagebackref=true,citecolor=green,bookmarks=false,hypertexnames=true]{hyperref}

\definecolor{darkgreen}{RGB}{0, 150, 0}
\definecolor{darkred}{RGB}{200, 0, 0}

\newcommand{\lone}{$L_1$\xspace}
\newcommand{\ltwo}{$L_2$\xspace}

\newcommand{\ch}{{\color{darkgreen} \ding{51}}}
\newcommand{\xm}{{\color{darkred} \ding{55}}}

\newcommand{\prob}[1]{p\left( #1 \right)}
\DeclarePairedDelimiter{\abs}{\lvert}{\rvert}

\def\wacvPaperID{0260} % Enter the WACV Paper ID here
\wacvfinalcopy % *** Uncomment this line for the final submission
\begin{document}
%%%%%%%%% TITLE
\title{SynDistNet: Self-Supervised Monocular Fisheye Camera Distance Estimation\\ Synergized with Semantic Segmentation for Autonomous Driving}

\author{Varun Ravi Kumar$^{1,4}$
\quad Marvin Klingner$^{3}$
\quad Senthil Yogamani$^{2}$\\ 
\quad Stefan Milz$^{4}$
\quad Tim Fingscheidt$^{3}$
\quad Patrick Mäder$^{4}$\\
$^{1}$Valeo DAR Kronach, Germany\quad
$^{2}$Valeo Vision Systems, Ireland\quad \\
$^{3}$Technische Universität Braunschweig, Germany\quad 
$^{4}$Technische Universität Ilmenau, Germany\\
}

\maketitle
\thispagestyle{empty}
%%%%%%%%% ABSTRACT
\begin{abstract}
State-of-the-art self-supervised learning approaches for monocular depth estimation usually suffer from scale ambiguity. They do not generalize well when applied on distance estimation for complex projection models such as in fisheye and omnidirectional cameras. This paper introduces a novel multi-task learning strategy to improve self-supervised monocular distance estimation on fisheye and pinhole camera images. Our contribution to this work is threefold: Firstly, we introduce a novel distance estimation network architecture using a self-attention based encoder coupled with robust semantic feature guidance to the decoder that can be trained in a one-stage fashion. Secondly, we integrate a generalized robust loss function, which improves performance significantly while removing the need for hyperparameter tuning with the reprojection loss. Finally, we reduce the artifacts caused by dynamic objects violating static world assumptions using a semantic masking strategy. We significantly improve upon the RMSE of previous work on fisheye by 25\% reduction in RMSE. As there is little work on fisheye cameras, we evaluated the proposed method on KITTI using a pinhole model. We achieved state-of-the-art performance among self-supervised methods without requiring an external scale estimation.
\end{abstract}
% -------------------------------------------------
% Introduction
\section{Introduction}

\begin{figure}[t]
  \captionsetup{singlelinecheck=false, font=small,  belowskip=-10pt}
  \centering
    \includegraphics[width=1.0\linewidth]{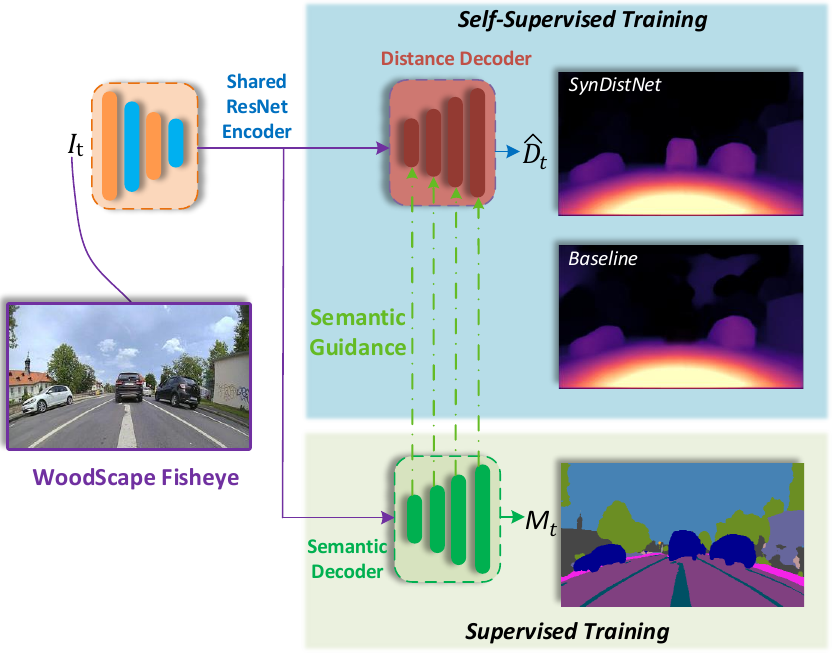}
    \caption{\textbf{Overview over the joint prediction of distance $\hat{D}_t$ and semantic segmentation $M_t$ from a single input image $I_t$.} Compared to previous approaches, our semantically guided distance estimation produces sharper depth edges and reasonable distance estimates for dynamic objects.
    }
    \label{fig:general_concept}
\end{figure}

Depth estimation plays a vital role in 3D geometry perception of a scene in various application domains such as virtual reality and autonomous driving. As LiDAR-based depth perception is sparse and costly, image-based methods are of significant interest in perception systems regarding coverage density and redundancy. Here, current state-of-the-art approaches do rely on neural networks \cite{Fu2018, Zhang2019b}, which can even be trained in an entirely self-supervised fashion from sequential images \cite{zhou2017unsupervised}, giving a clear advantage in terms of applicability to arbitrary data domains over supervised approaches.\par
While most academic works focus on pinhole cameras \cite{zhou2017unsupervised, garg2016unsupervised, mahjourian2018unsupervised, Godard2019}, many real-world applications rely on more advanced camera geometries as, e.g., fisheye camera images. There is little work on visual perception tasks on fisheye cameras \cite{uvrivcavr2019soilingnet, yahiaoui2019fisheyemodnet, kumar2018monocular, dahal2019deeptrailerassist, tripathi2020trained, uricar2019desoiling}. With this observation, in this paper, we present our proposed improvements not only on pinhole camera images but also on fisheye camera images (cf. Ravi Kumar \etal~\cite{kumar2020fisheyedistancenet, kumar2020unrectdepthnet}). We show that our self-supervised distance estimation (a generalization of depth estimation) works for both considered camera geometries. Our first contribution in that sense is the novel application of a general and robust loss function proposed by~\cite{barron2019general} to the task of self-supervised distance estimation, which replaces the de facto standard of an \lone loss function used in previous approaches~\cite{Casser2019, Godard2019, Guizilini2020a, kumar2020unrectdepthnet, kumar2020fisheyedistancenet}.\par
As the distance predictions are still imperfect due to the monocular cues such as occlusion, blur, haze, and different lighting conditions and the dynamic objects during the self-supervised optimizations between consecutive frames. Many approaches consider different scene understanding modalities, such as segmentation~\cite{Meng2019a, Guizilini2020, Ranjan2019} or optical flow~\cite{Yang2018c, Chen2019b} within multi-task learning to guide and improve the distance estimation. As optical flow is usually also predicted in a self-supervised fashion~\cite{Liu2019} it is therefore subject to similar limitations as the self-supervised distance estimation, which is why we focus on the joint learning of self-supervised distance estimation and semantic segmentation.\par
In this context, we propose a novel architecture for the joint learning of self-supervised distance estimation and semantic segmentation, which introduces a significant change compared to earlier works~\cite{Guizilini2020}. We first propose the novel application of self-attention layers in the ResNet encoder used for distance estimation. We also employ pixel adaptive convolutions within the decoder for robust semantic feature guidance, as proposed by~\cite{Guizilini2020}. However, we train the semantic segmentation simultaneously, which introduces a more favorable one-stage training than other approaches relying on pre-trained models~\cite{Casser2019, Casser2019a, Guizilini2020, Meng2019a}.\par
As depicted in Fig.\ \ref{fig:general_concept}, dynamic objects induce a lot of unfavorable artifacts and hinder the photometric loss during the training, which results in infinite distance predictions, e.g., due to their violation of the static world assumption. Therefore, we use the segmentation masks to apply a simple semantic masking technique, based on the temporal consistency of consecutive frames, which delivers significantly improved results, e.g., concerning the infinite depth problem of objects, moving at the same speed as the ego-camera. Previous approaches~\cite{Luo2019a, Ranjan2019, Yang2018c} did predict these motion masks only implicitly as part of the projection model and therefore were limited to the projection model's fidelity.\par
Our contributions are the following: Firstly, we introduce a novel architecture for the learning of self-supervised distance estimation synergized with semantic segmentation. Secondly, we improve the self-supervised distance estimation by a general and robust loss function. Thirdly, we propose a solution for the dynamic object impact on self-supervised distance estimation by using semantic-guidance. We show the effectiveness of our approach both on pinhole and fisheye camera datasets and present state-of-the-art results for both image types.\par
% Related Work
\section{Related Work}
\label{sec:related_work}

In this section, we first provide an overview of self-supervised depth/distance estimation approaches. Afterward, we discuss their combination with other tasks in multi-task learning settings and particular methods utilizing semantic guidance.\par
% -------------------------------------------------
\textbf{\textit{Self-Supervised Depth Estimation}}
%After Eigen \etal~\cite{Eigen_14} successfully proved that fully convolutional neural networks are capable of predicting depth from single images, the approaches of 
Garg \etal~\cite{garg2016unsupervised}, and Zhou \etal~\cite{zhou2017unsupervised} showed that it is possible to train networks in a self-supervised fashion by modeling depth as part of a geometric projection between stereo images and sequential images, respectively. The initial concept has been extended by considering improved loss functions~\cite{Aleotti2018, Godard2017, mahjourian2018unsupervised, Godard2019}, the application of generative adversarial networks (GANs) \cite{Aleotti2018, Kumar2018a, Pilzer2018} or generated proxy labels from traditional stereo algorithms \cite{Tosi2019}, or synthetic data~\cite{Bozorgtabar2019}. Other approaches proposed to use specialized architectures for self-supervised depth estimation \cite{Guizilini2020a,Wang2018e,Zhou2019}, they apply teacher-student learning \cite{Pilzer2019} to use test-time refinement strategies~\cite{Casser2019,Casser2019a}, to employ recurrent neural networks \cite{Wang2019,Zhang2019c}, or to predict the camera parameters \cite{Gordon2019} to enable training across images from different cameras.\par
A recent approach by Ravi Kumar \etal~\cite{kumar2020fisheyedistancenet, kumar2020unrectdepthnet} presents a successful proof of concept for the application of self-supervised depth estimation methods on the task of distance estimation from fish-eye camera images, which is used as a baseline during this work. Recent approaches also investigated the application of self-supervised depth estimation to $360^\circ$ images~\cite{Wang2020b, Jin2020}. However, apart from these works, the application of self-supervised depth estimation to more advanced geometries, such as fish-eye camera images, has not been investigated extensively, yet.\par
% FIG -------------------------------------------------
\begin{figure*}[t]
  \captionsetup{singlelinecheck=false, font=small, belowskip=-10pt}
  \centering
    \includegraphics[width=0.7\textwidth]{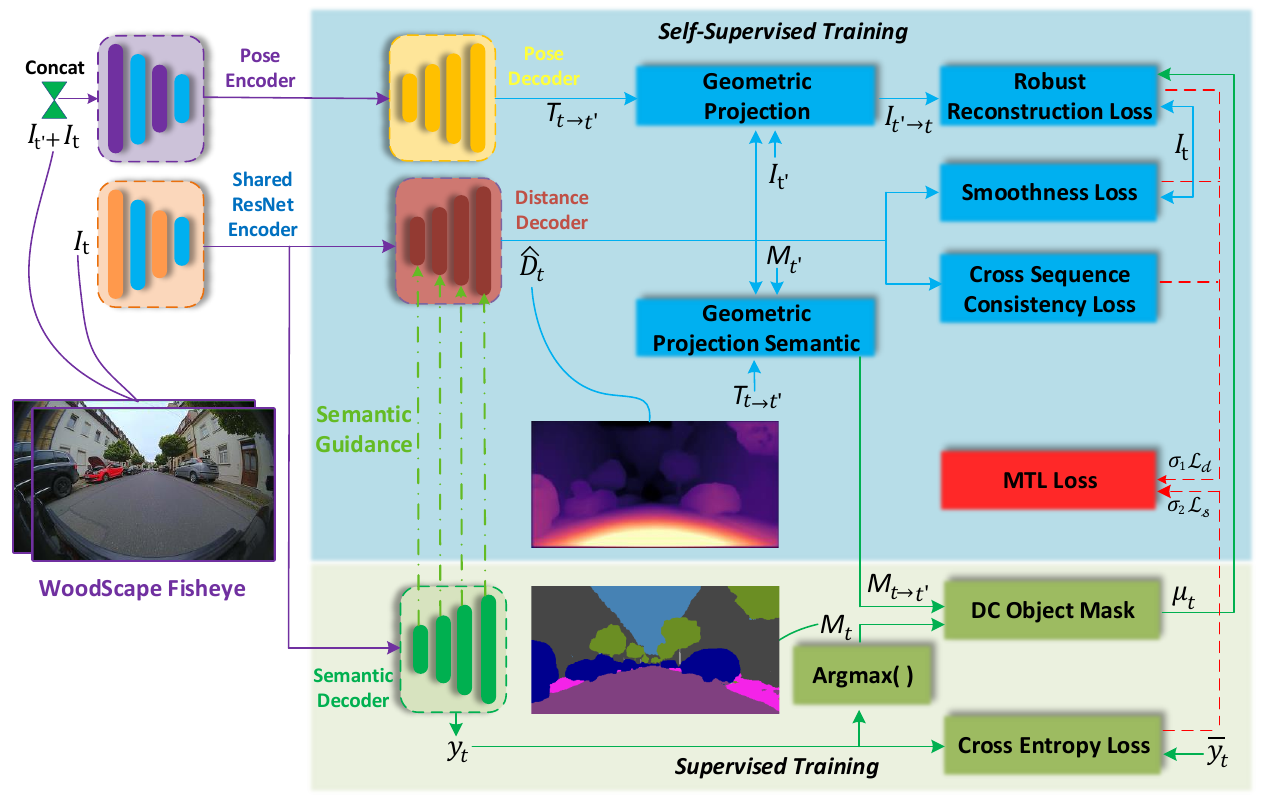}
    \caption{\textbf{Overview of our proposed framework} for the joint prediction of distance and semantic segmentation. The upper part (blue blocks) describes the single steps for the depth estimation, while the green blocks describe the single steps needed for the prediction of the semantic segmentation. Both tasks are optimized inside a multi-task network by using the weighted total loss described in Eq. \ref{eq:mtl_loss}.}
    \label{fig:mtl_pipeline}
\end{figure*}
%-------------------------------------------------
\textbf{\textit{Multi-Task Learning}}
In contrast to letting a network predict one single task, it is also possible to train a network to predict several tasks at once, which has been shown to improve tasks such as, \eg, semantic segmentation, \cite{Klingner2020, klingner2020selfsupervised, sistu2019neurall, chennupati2019multinet++, chennupati2019auxnet}, domain adaptation~\cite{Bolte2019a,Ochs2019,Zhao2019}, instance segmentation: \cite{Kirillov2019} and depth estimation \cite{eigen2015predicting, Kendall2018, Wang2020a}. While initial works did weigh losses \cite{eigen2015predicting} or gradients \cite{Ganin2015} by an empirical factor, current approaches can estimate this scale factor automatically \cite{Kendall2018, Chen2018d}. We adopt the uncertainty-based task weighting of Kendall \etal \cite{Kendall2018}.\par
Many recent approaches aim to integrate optical flow into the self-supervised depth estimation training, as this additional task can also be trained in a self-supervised fashion~\cite{ Liu2019, Rhe2017}. In these approaches, both tasks are predicted simultaneously. Then losses are applied to enforce cross-task consistency \cite{Liu2019a, Luo2019a, Wang2019c, Yang2018c}, to enforce known geometric constraints \cite{Chen2019b, Ranjan2019}, or to induce a modified reconstruction of the warped image \cite{Chen2019b, Yin2018}. Although the typical approach is to compensate using optical flow, we propose an alternative method to use semantic segmentation instead for two reasons. Firstly, semantic segmentation is a mature and common task in autonomous driving, which can be leveraged. Second of all, optical flow is computationally more complex and harder to validate because of difficulties in obtaining ground truth.\par
%-------------------------------------------------
\textbf{\textit{Semantically-Guided Depth Estimation}}
Several recent approaches also used semantic or instance segmentation techniques to identify moving objects and handle them accordingly inside the photometric loss \cite{Casser2019, Casser2019a, Meng2019a, Vijayanarasimhan2017, Guizilini2020}. To this end, the segmentation masks are either given as an additional input to the network \cite{Meng2019a, Guizilini2020} or used to predict poses for each object separately between two consecutive frames \cite{Casser2019, Casser2019a, Vijayanarasimhan2017} and apply a separate rigid transformation for each object. Avoiding an unfavorable two-step (pre)training procedure, other approaches in \cite{Chen2019a, Novosel2019, Yang2018b, Zhu2020} train both tasks in one multi-task network simultaneously, improving the performance by cross-task guidance between these two facets of scene understanding. Moreover, the segmentation masks can be projected between frames to enforce semantic consistency \cite{Chen2019a, Yang2018b}, or the edges can be enforced to appear in similar regions in both predictions \cite{Chen2019a, Zhu2020}. In this work, we propose to use this warping to discover frames with moving objects and learn their depth from these frames by applying a simple semantic masking technique. We also propose a novel self-attention-based encoder and semantic features guidance to the decoder using pixel-adaptive convolutions as in~\cite{Guizilini2020}. We can apply a one-stage training by this simple change, removing the need to pretrain a semantic segmentation network.
% -------------------------------------------------
% Method
% -------------------------------------------------
\section{Multi-Task Learning Framework}
\label{sec:method}

In this section, we describe our framework for the multi-task learning of distance estimation and semantic segmentation. We first state how we train the tasks individually and how they are trained in a synergized fashion.
% -------------------------------------------------
\subsection{Self-Supervised Distance Estimation Baseline}

Our self-supervised depth and distance estimation is developed within a self-supervised monocular structure-from-motion (SfM) framework which requires two networks aiming at learning:
\begin{enumerate}[nosep]
\item a monocular depth/distance model $g_D: I_t \to \hat D_t$ predicting a scale-ambiguous depth or distance (the equivalent of depth for general image geometries) $\hat D_t = g_D(I_t(ij))$ per pixel $ij$ in the target image $I_t$; and
\item an ego-motion predictor $g_T: (I_t, I_{t^\prime}) \to T_{t \to t^\prime}$ predicting a set of 6 degrees of freedom which implement a rigid transformation $T_{t \rightarrow t'} \in \text{SE(3)}$, between the target image $I_t$ and the set of reference images $I_{t^\prime}$.
Typically, $t' \in \{t+1, t-1\}$, \ie the frames $I_{t-1}$ and $I_{t+1}$ are used as reference images, although using a larger window is possible.
\end{enumerate}
In the following part, we will describe our different loss contributions in the context of fisheye camera images.\par
% -------------------------------------------------
\textbf{\textit{Total Self-Supervised Objective Loss}}
View synthesis is performed by incorporating the projection functions from FisheyeDistanceNet~\cite{kumar2020fisheyedistancenet}, and the same protocols are used to train the distance and pose estimation networks simultaneously. Our self-supervised objective loss consists of a reconstruction matching term $\mathcal{L}_r$ that is calculated between the reconstructed $\hat{I}_{t'\to t}$ and original $I_t$ target images, and an inverse depth or distance regularization term $\mathcal{L}_s$ introduced in \cite{Godard2017} that ensures edge-aware smoothing in the distance estimates $\hat{D}_t$. Finally, we apply a cross-sequence distance consistency loss $\mathcal{L}_{dc}$ derived from the chain of frames in the training sequence and the scale recovery technique from \cite{kumar2020fisheyedistancenet}. The final objective loss $\mathcal{L}_{tot}$ is averaged per pixel, scale and image batch, and is defined as: 
\begin{align}
    \mathcal{L}_{tot} = \mathcal{L}_r(I_t,\hat{I}_{t'\to t}) + \beta~\mathcal{L}_s(\hat{D}_t) + \gamma~\mathcal{L}_{dc}(\hat{D}_t,\hat{D}_{t'})
    \label{eq:overall-loss}
\end{align}
where $\beta$ and $\gamma$ are weight terms between the distance regularization $\mathcal{L}_{s}$ and the cross-sequence distance consistency $\mathcal{L}_{dc}$ losses, respectively.\par

% -------------------------------------------------
\textbf{\textit{Image Reconstruction Loss}}
\label{sec:reconstruction loss}
Most state-of-the-art self-supervised depth estimation methods use heuristic loss functions. However, the optimal choice of a loss function is not well defined theoretically. In this section, we emphasize the need for exploration of a better photometric loss function and explore a more generic robust loss function.\par
Following previous works~\cite{kumar2020fisheyedistancenet, Godard2017, Godard2019, zhao2016loss, Guizilini2020a}, the image reconstruction loss between the target image $I_t$ and the reconstructed target image $\hat I_{t' \to t}$ is calculated using the \lone pixel-wise loss term combined with Structural Similarity (SSIM)~\cite{wang2004image}
\begin{align}
  \label{eq:loss-photo}
  \tilde{\mathcal{L}}_{r}(I_t,\hat I_{t' \to t}) &= \omega~\frac{1 - \text{SSIM}(I_t,\hat I_{t' \to t})}{2} \nonumber \\
  &\quad+ (1-\omega)~ \left\| (I_t - \hat I_{t' \to t}) \right\|
\end{align}

where $\omega=0.85$ is a weighting factor between both loss terms. The final per-pixel minimum reconstruction loss $\mathcal{L}_{r}$ \cite{Godard2019} is then calculated over all the source images
\begin{align}
    \mathcal{L}_{r} &= \min_{t^\prime \in \{t+1,t-1\}} \tilde{\mathcal{L}}_r(I_t, \hat I_{t' \to t})
\end{align}\par
We also incorporate the insights introduced in~\cite{Godard2019}, namely auto-masking, which mitigates the impact of static pixels by removing those with unchanging appearance between frames and inverse depth map upsampling which helps to removes texture-copy artifacts and holes in low-texture regions.

% -------------------------------------------------
\subsection{Semantic Segmentation Baseline}
We define semantic segmentation as the task of assigning a pixel-wise label mask $M_{t}$ to an input image $I_t$, i.e. the same input as for distance estimation from a single image. Each pixel gets assigned a class label $s\in\mathcal{S} =  \left\lbrace1,2,...,S\right\rbrace$ from the set of classes $\mathcal{S}$. In a supervised way, the network predicts a posterior probability $Y_t$ that a pixel belongs to a class $s\in \mathcal{S}$, which is then compared to the one-hot encoded ground truth labels $\overline{Y}_{t}$ inside the cross-entropy loss
\begin{equation}
\mathcal{L}_{ce} = -\sum_{s \in\mathcal{S}} \overline{Y}_{t,s} \cdot \log\left(Y_{t,s}\right)
\label{eq:crossentropy_loss}
\end{equation}
the final segmentation mask $M_{t}$ is then obtained by applying a pixel-wise $\operatorname{argmax}$ operation on the posterior probabilities $Y_{t,s}$. Note that we also use unrectified fisheye camera images, for which the segmentation task can however still be applied as shown in this work.\par
\begin{figure}[t]
    \captionsetup{belowskip=-8pt, font= small, singlelinecheck=false}
	\centering
	\subfloat[][Image \label{fig:mask_analysis_a}]{\includegraphics[width=0.48\linewidth]{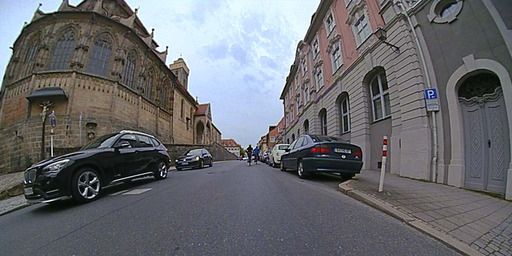}}\;\;
	\subfloat[][Segmentation \label{fig:mask_analysis_b}]{\includegraphics[width=0.48\linewidth]{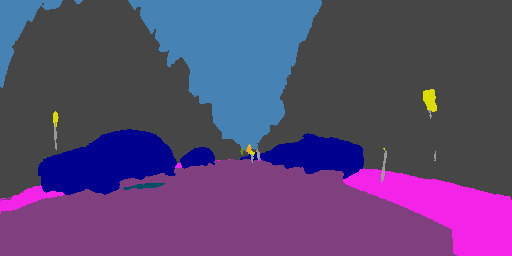}}\\
	\vspace{-3mm}
	\subfloat[][Projected image \label{fig:mask_analysis_c}]{\includegraphics[width=0.48\linewidth]{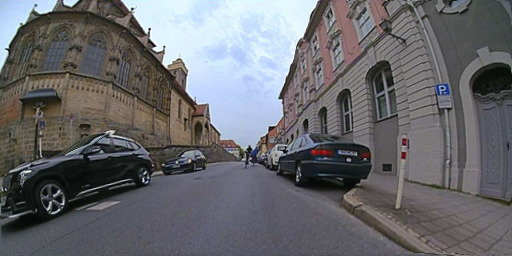}}\;\;
	\subfloat[][Projected segmentation \label{fig:mask_analysis_d}]{\includegraphics[width=0.48\linewidth]{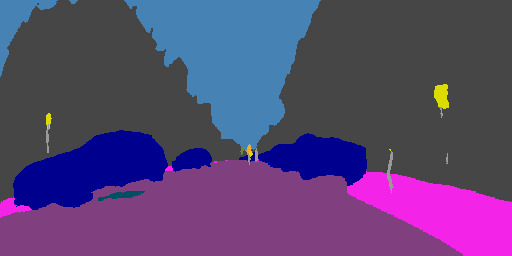}}\\
	\vspace{-3mm}
	\subfloat[][Photometric error \label{fig:mask_analysis_e}]{\includegraphics[width=0.48\linewidth]{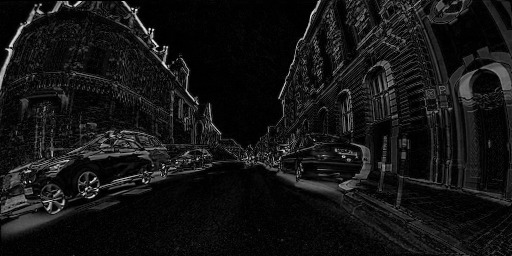}} \;\;
	\subfloat[][Dynamic object mask \label{fig:mask_analysis_f}]{\includegraphics[width=0.48\linewidth]{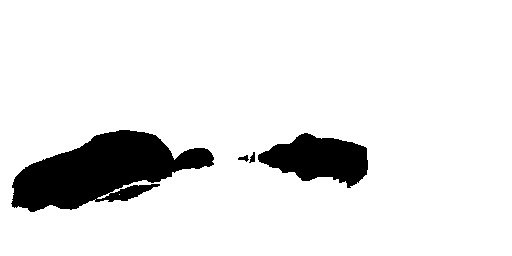}}\\
	\vspace{-3mm}
	\subfloat[][Distance Estimate \label{fig:mask_analysis_g}]{\includegraphics[width=0.48\linewidth]{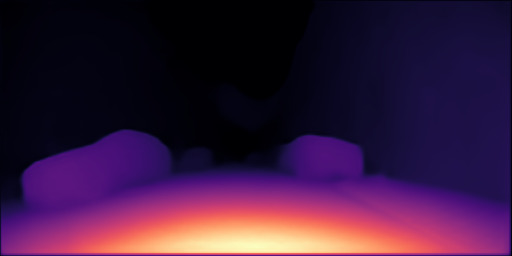}} \;\;
	\subfloat[][ Mask (f) applied on (e) \label{fig:mask_analysis_h}]{\includegraphics[width=0.48\linewidth]{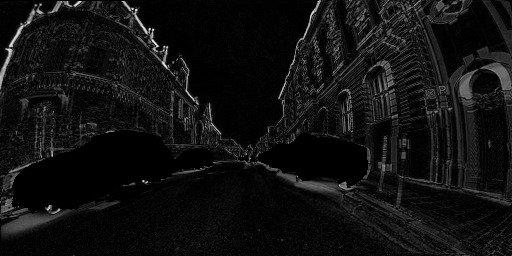}}\\
	\caption{Application of our semantic masking methods, to handle potentially dynamic objects. The dynamic objects inside the segmentation masks from consecutive frames in (b) and (d) are accumulated to a dynamic object mask, which is used to mask the photometric error (e), as shown in (h).}
	\label{fig:mask_analysis}
\end{figure} 

% -------------------------------------------------
\subsection{Robust Reconstruction Loss for Distance Estimation}

Towards developing a more robust loss function, we introduce the common notion of a per-pixel regression $\rho$ in the context of depth estimation, which is given by 
\begin{align}
    \rho \left( \xi \right) = \rho \left( \hat I_{t' \to t}-I_t \right)
\end{align}
while this general loss function can be implemented by a simple \lone loss as in the second term of Eq.~\ref{eq:loss-photo}, recently, a general and more robust loss function is proposed by Barron~\cite{barron2019general}, which we use to replace the \lone term in Eq.~\ref{eq:loss-photo}. This function is a generalization of many common losses such as the \lone, \ltwo, Geman-McClure, Welsch/Leclerc, Cauchy/Lorentzian and Charbonnier loss functions. In this loss, robustness is introduced as a continuous parameter and it can be optimized within the loss function to improve the performance of regression tasks. This robust loss function $\rho_{\mathrm{rob}}$ is given by:
\begin{align}
    \label{equ:robust_loss}
    \rho_{\mathrm{rob}}\left(\xi\right) = \frac{\abs{\alpha - 2}}{\alpha} \left( \left( {\frac{\left( \sfrac{\xi}{c} \right)^2}{\abs{\alpha - 2}}} + 1 \right)^{\sfrac{\alpha}{2}} - 1 \right)
\end{align}
The free parameters $\alpha$, and $c$ in this loss can be automatically adapted to the particular problem via a data-driven optimization, as described in~\cite{barron2019general}.

% -------------------------------------------------
%%%%%COMMENTED
\begin{comment}
To induce $\alpha$ as a trainable parameter Barron~\cite{barron2019general} encapsulates the loss into a probability density function given by:
\begin{align}
    \prob{x \;|\; \mu, \alpha, c} &= {1 \frac {c}{\partition{\alpha}}} \exp \left( -\rho_{\mathrm{rob}}\left(x - \mu, \alpha, c\right) \right) \\
    \partition{\alpha} &= \int_{-\infty}^{\infty} \exp \left( -\rho_{\mathrm{rob}}\left(x, \alpha, 1\right) \right)
\end{align}
where $\prob{x \;|\; \mu, \alpha, c}$ is only defined if $\alpha \geq 0$, as $\partition{\alpha}$ is divergent when
$\alpha < 0$. Then the optimization function boils down to:
\begin{align}
    \arg \min_{\theta,\alpha} -log(p(x|\alpha) = \rho \left( x,\alpha \right ) + log(Z(\alpha))
\end{align}
where $log(Z(\alpha))$ is an analytical function which is approximated with a cubic spline function. $\partition{\alpha}$ is an important factor in the loss function as it reduces the cost of outliers. The main properties of the loss are: \textit{i)} It is monotonic with respect to its inputs $\abs{x}$ and $\alpha$ which is useful for graduated non-convexity. \textit{ii)} It is smooth with respect to its inputs $x$ and $\alpha$ (i.e in $C^\infty$). \textit{iii)} It has bounded first and second derivatives (no exploding gradients and easier pre-conditioning).
\end{comment}

% -------------------------------------------------
\subsection{Dealing With Dynamic Objects}

Typically, the assumed static world model for projections between image frames is violated by the appearance of dynamic objects. Thereby, we use the segmentation masks to exclude \textit{moving} potentially dynamic objects while \textit{non-moving} dynamic object should still contribute.\par 
In order to implement this, we aim at defining a pixel-wise mask $\mu_t$, which contains a $0$, if a pixel belongs to a dynamic object from the current frame $I_t$, or to a wrongfully projected dynamic object from the reconstructed frames $\hat{I}_{t'\to t}$, and a $1$ otherwise. For calculation of the mask, we start by predicting a semantic segmentation mask $M_t$ which corresponds to image $I_t$ and also segmentation masks $M_{t'}$ for all images $I_{t'}$. Then we use the same projections as for the images and warp the segmentation masks (using nearest neighbour instead of bilinear sampling), yielding projected segmentation masks $M_{t' \to t}$. Then, also defining the set of dynamic object classes $\mathcal{S}_{\mathrm{DC}} \subset \mathcal{S}$ we can define $\mu_t$ by its pixel-wise elements at pixel location $ij$:
\begin{equation}
\!\!\mu_{t, ij} = 
\!\left\{
\begin{array}{l}
1 ,\; M_{t, ij} \notin \mathcal{S}_{\mathrm{DC}}\; \land \; M_{t'\rightarrow t,ij} \notin \mathcal{S}_{\mathrm{DC}} \\
0,\; \mathrm{else} \\
\end{array}
\right.
\label{eq:semantic_mask}
\end{equation}

The mask is then applied pixel-wise on the reconstruction loss defined in Eq.~\ref{eq:loss-photo}, in order to mask out dynamic objects. However, as we only want to mask out \textit{moving} DC-objects, we detect them using the consistency of the target segmentation mask and the projected segmentation mask to judge whether dynamic objects are moving between consecutive frames (e.g., we intend to learn the depth of dynamic objects from parking cars, but not from driving ones). With this measure, we apply the dynamic object mask $\mu_t$ only to an imposed fraction $\epsilon$ of images, in which the objects are detected as mostly moving.

% -------------------------------------------------
\subsection{Joint Optimization}

We incorporate the task weighting approach by Kendall \etal~\cite{Kendall2018}; we weigh our distance estimation and semantic segmentation loss terms for multi-task learning, which enforces homoscedastic (task) uncertainty. It is proven to be effective in weighing the losses from Eq.~\ref{eq:overall-loss} and Eq.~\ref{eq:crossentropy_loss} by:
\begin{align}
    \label{eq:mtl_loss}
    \frac{1}{2 \sigma_1^2} \mathcal{L}_{tot} + \frac{1}{2 \sigma_2^2} \mathcal{L}_{ce} + \log (1 + \sigma_1) + \log (1+ \sigma_2)
\end{align}
Homoscedastic uncertainty does not change with varying input data and is task-specific. We, therefore, learn this uncertainty and use it to down weigh each task. Increasing the noise parameter $\sigma$ reduces the weight for the respective task. Furthermore, $\sigma$ is a learnable parameter; the objective optimizes a more substantial uncertainty that should lead to a smaller contribution of the task's loss to the total loss. In this case, the different scales from the distance and semantic segmentation are weighed accordingly. The noise parameter $\sigma_1$ tied to distance estimation is quite low compared to $\sigma_2$ of semantic segmentation, and the convergence occurs accordingly. Higher homoscedastic uncertainty leads to a lower impact of the task's network weight update. It is important to note that this technique is not limited to the joint learning of distance estimation and semantic segmentation, but can also be applied to more tasks and arbitrary camera geometries.
% -------------------------------------------------
% Network Details
% -------------------------------------------------
\section{Network Architecture}
\label{sec:network_details}

In this section, we will describe our novel architecture for self-supervised distance estimation utilizing semantic guidance. The baseline from \cite{kumar2020fisheyedistancenet} used deformable convolutions to model the fisheye geometry to incorporate the distortion and improve the distance estimation accuracy. In this work, we introduce a self-attention based encoder to handle the view synthesis and a semantically guided decoder, which can be trained in a one-stage fashion.
% -------------------------------------------------
\begin{figure}[t]
  \captionsetup{singlelinecheck=false, font=small, belowskip=-10pt}
  \centering
    \includegraphics[width=\columnwidth]{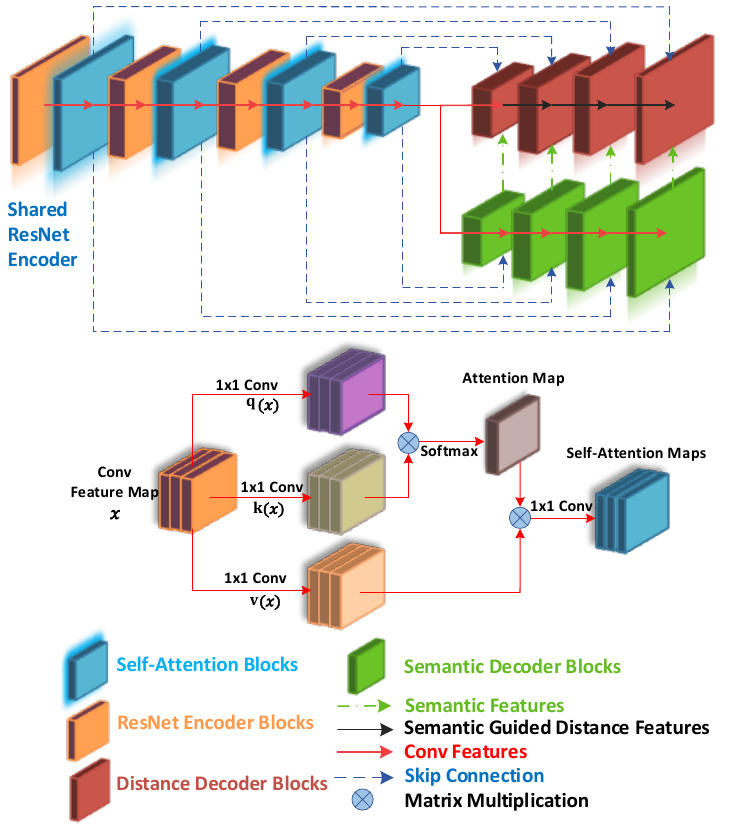}
    \caption{\textbf{Visualization of our proposed network architecture} to semantically guide the depth estimation. We utilize a self-attention based encoder and a semantically guided decoder using pixel-adaptive convolutions.}
    \label{fig:model_arch}
\end{figure}
% -------------------------------------------------
\subsection{Self-Attention Encoder}

Previous depth estimation networks~\cite{Godard2019, zhou2017unsupervised} use normal
convolutions for capturing the local information in an image, but the convolutions' receptive field is relatively small.
Inspired by~\cite{ramachandran2019stand}, who took self-attention in CNNs even further by using stand-alone self-attention blocks instead of only enhancing convolutional layers. The authors present a self-attention layer which may replace convolution while reducing the number of parameters. Similar to a convolution, given a pixel $x_{ij} \in \mathbb{R}^{d_{in}}$ inside a feature map, the local region of pixels defined by positions ${a b} \in \mathcal{N}_k(ij)$ with spatial extent $k$ centered around $x_{ij}$ are extracted initially which is referred to as a memory block. For every memory block, the single-headed attention for computing the pixel output $z_{ij} \in \mathbb{R}^{d_{out}}$ is then calculated by:
\begin{align}
\label{eq:attention}
z_{ij} = \sum_{\mathclap{ab \in \, \mathcal{N}_k(ij)}}
         \texttt{softmax}_{a b}\left(q_{i j}^\top k_{a b} \right) v_{a b}
\end{align}  
where $q_{ij} = W_Q x_{ij}$ are the \emph{queries}, \emph{keys} $k_{ab} = W_K x_{ab}$, and \emph{values} $v_{ab} = W_V x_{ab}$ are linear transformations of the pixel in position $ij$ and the neighborhood pixels. The learned transformations are denoted by the matrices W. $\texttt{softmax}_{a b}$ defines a softmax applied to all logits computed in the neighborhood of $ij$. $W_Q, W_K, W_V \in \mathbb{R}^{d_{out} \times d_{in}}$ are trainable transformation weights. There exists an issue in the above-discussed approach, as there is no positional information encoded in the attention block. Thus the Eq.~\ref{eq:attention} is invariant to permutations of the individual pixels. For perception tasks, it is typically helpful to consider spatial information in the pixel domain. For example, the detection of a pedestrian is composed of spotting faces and legs in a proper relative localization. The main advantage of using self-attention layers in the encoder is that it induces a synergy between geometric and semantic features for distance estimation and semantic segmentation tasks. In~\cite{vaswani2017attention} sinusoidal embeddings are used to produce the absolute positional information. Following~\cite{ramachandran2019stand}, instead of attention with 2D relative position embeddings, we incorporate relative attention due to their better accuracy for computer vision tasks. The relative distances of the position $ij$ to every neighborhood pixel $(a,b)$ is calculated to obtain the relative embeddings. The calculated distances are split up into row and column distances $r_{a-i}$ and $r_{b-j}$ and the embeddings are concatenated to form $r_{a-i,b-j}$ and multiplied by the query $q_{ij}$ given by:
\begin{equation} \label{equation:standard-self-attention}
z_{ij} = \sum_{\mathclap{a b \in \, \mathcal{N}_k(i j)}}
         \texttt{softmax}_{a b}\left( q_{i j}^\top  k_{a b} + q_{i j}^\top r_{a-i,b-j} \right)  v_{a b}
\end{equation}
It ensures the weights calculated by the softmax function are modulated by both the relative distance and content of the key from the query. Instead of focusing on the whole feature map, the attention layer only focuses on the memory block.
% -------------------------------------------------
\subsection{Semantically-Guided Distance Decoder}

To address the limitations of regular convolutions, we follow the approaches of \cite{su2019pixel, Guizilini2020} in using pixel-adaptive convolutions for semantic guidance inside the distance estimation branch of the multi-task network. By this approach, we can break up the translation invariance of convolutions and incorporate spatially-specific information of the semantic segmentation branch.\par
To this end, as shown in Figure~\ref{fig:model_arch} we extract feature maps at different levels from the semantic segmentation branch of the multi-task network. These semantic feature maps are consequently used to guide the respective pixel-adaptive convolutional layer, following the formulation proposed in~\cite{su2019pixel} to process an input signal $x$ to be convolved:
\begin{equation}
x_{ij}' = \sum_{ab \in \mathcal{N}_k(i, j)} K(F_{ij},F_{ab}) W [r_{a-i,b-j}]x_{ab} + B
\end{equation}
where $\mathcal{N}_k(i, j)$ defines a $k \times k$ neighbourhood window around the pixel location $ij$ (distance $r_{a-i,b-j}$ between pixel locations), which is used as input to the convolution with weights $W$ (kernel size $k$), bias $B\in \mathbb{R}^1$ and kernel $K$, that is used in this case to calculate the correlation between the semantic guidance features $F \in \mathbb{R}^D$ from the segmentation network. We follow \cite{Guizilini2020} in using a Gaussian kernel:
\begin{equation}
   K(F_{ij},F_{ab}) = \exp\left(-\frac{1}{2} (F_{ij} - F_{ab})^T \Sigma_{ijab}^{-1} (F_{ij} - F_{ab})\right)
\end{equation}
with covariance matrix $\Sigma_{ijab}$ between features $F_{ij}$ and $F_{ab}$, which is chosen as a diagonal matrix $\sigma^2 \cdot\mathbf{1}^D$, where $\sigma$ represents a learnable parameter for each convolutional filter.\par
In this work, we use pixel-adaptive convolutions to produce \emph{semantic-aware distance features}, where the fixed information encoded in the semantic network is used to disambiguate geometric representations for the generation of multi-level depth features. Compared to previous approaches \cite{Casser2019, Guizilini2020}, we use features from our semantic segmentation branch that is trained simultaneously with the distance estimation branch introducing a more favorable one-stage training.
% -------------------------------------------------
% Experiments
% -------------------------------------------------
\section{Experimental Evaluation}
\label{sec:experiments}

\vspace{-5pt}
\begin{table}[b]
  \captionsetup{belowskip=-8pt, font= small, singlelinecheck=false}
  \centering
    \begin{adjustbox}{width=\columnwidth}
      \begin{tabular}{l|c|c}
      \toprule
      \multicolumn{1}{l|}{\textbf{Model}} 
      & \multicolumn{1}{c|}{\cellcolor[HTML]{00b050}\begin{tabular}[c]{@{}c@{}}\textit{Segmentation} \\ (mIOU)\end{tabular}} & \multicolumn{1}{c}{\cellcolor[HTML]{00b0f0}\begin{tabular}[c]{@{}c@{}} \textit{Distance} \\ (RMSE) \end{tabular}} \\ 
      \midrule 
      Segmentation only baseline     & 76.8         & \xm     \\ 
      Distance only baseline         & \xm          & 2.316 \\ 
      MTL baseline                   & 78.3         & 2.128 \\ 
      MTL with synergy (SynDistNet) & \textbf{81.5} & \textbf{1.714} \\ 
      \bottomrule
    \end{tabular}
\end{adjustbox}
\vspace{-5pt}
\caption{\textbf{Multi-task learning (MTL) ablation results} on the WoodScape dataset using a ResNet18 encoder.}
\label{tab:mtl}
\end{table}
% -------------------------------------------------
\begin{table*}[htpb]
\captionsetup{belowskip=-6pt, font= small, singlelinecheck=false}
  \centering{
  \small
  \setlength{\tabcolsep}{0.3em}
  \begin{tabular}{l|c|c|c|c|cccc|ccc}
    \toprule
      \multirow{2}{*}{
       \textbf{Network} } & 
       \multirow{2}{*}{\emph{RL}} &
       \multirow{2}{*}{\emph{Self-Attn}} & 
       \multirow{2}{*}{\emph{SEM}} & 
       \multirow{2}{*}{\emph{Mask}} & 
       \multicolumn{4}{c|}{\cellcolor[HTML]{5880ab}Lower is Better} & 
       \multicolumn{3}{c}{\cellcolor[HTML]{e8715b}Higher is Better} \\
        & & &  &     & \cellcolor[HTML]{5880ab}Abs Rel & \cellcolor[HTML]{5880ab}Sq Rel & \cellcolor[HTML]{5880ab}RMSE & \cellcolor[HTML]{5880ab}RMSE$_{log}$ & \cellcolor[HTML]{e8715b}$\delta < 1.25$ & \cellcolor[HTML]{e8715b}$\delta < 1.25^2$ & \cellcolor[HTML]{e8715b}$\delta < 1.25^3$ \\
       \toprule
       FisheyeDistanceNet \cite{kumar2020fisheyedistancenet} & \xm & \xm & \xm & \xm & 0.152 & 0.768 & 2.723 & 0.210 & 0.812 & 0.954 & 0.974 \\
       \midrule
        \multirow{6}{*}{SynDistNet (ResNet-18)}
        & \ch & \xm & \xm & \xm & 0.142 & 0.537 & 2.316 & 0.179 & 0.878 & 0.971 & 0.985 \\
        & \ch & \xm & \xm & \ch & 0.133 & 0.491 & 2.264 & 0.168 & 0.868 & 0.976 & 0.988 \\
        & \ch & \ch & \xm & \ch & 0.121 & 0.429 & 2.128 & 0.155 & 0.875 & 0.980 & 0.990 \\
        & \ch & \ch & \ch & \xm & 0.105 & 0.396 & 1.976 & 0.143 & 0.878 & 0.982 & 0.992 \\
        & \ch & \ch & \ch & \ch & \textbf{0.076} & \textbf{0.368} & \textbf{1.714} & \textbf{0.127} & \textbf{0.891} & \textbf{0.988} & \textbf{0.994} \\
        \midrule
        \multirow{6}{*}{SynDistNet (ResNet-50)}
        & \ch & \xm & \xm & \xm & 0.138 & 0.540 & 2.279 & 0.177 & 0.880 & 0.973 & 0.986 \\
        & \ch & \xm & \xm & \ch & 0.127 & 0.485 & 2.204 & 0.166 & 0.881 & 0.975 & 0.989 \\
        & \ch & \ch & \xm & \ch & 0.115 & 0.413 & 2.028 & 0.148 & 0.876 & 0.983 & 0.992 \\
        & \ch & \ch & \ch & \xm & 0.102 & 0.387 & 1.856 & 0.135 & 0.884 & 0.985 & 0.994 \\
        & \ch & \ch & \ch & \ch & \textbf{0.068} & \textbf{0.352} & \textbf{1.668} & \textbf{0.121} & \textbf{0.895} & \textbf{0.990} & \textbf{0.996} \\
   \bottomrule
  \end{tabular}
  }
  \vspace{-5pt}
\caption{\textcolor{black}{
  \textbf{Ablative analysis} showing the effect of each of our contributions using the Fisheye WoodScape dataset~\cite{yogamani2019woodscape}. The input resolution is $512 \times 256$ pixels and distances are capped at $40\,m$. We start with FisheyeDistanceNet \cite{kumar2020fisheyedistancenet} baseline and incrementally add robust loss (RL), self-attention based encoder (Self-Attn), semantically-guided decoder (SEM) and dynamic object masking (Mask). }
  }
\label{table:table_ablation}
\end{table*}
% -------------------------------------------------
% TABLE: Qualitative results on the fisheye WoodScape dataset for dynamic mask and infinite depth issue
% -------------------------------------------------
\begin{table*}[htpb]
 \captionsetup{belowskip=-10pt, font= small, singlelinecheck=false}
  \small
    \begin{center}
	  \begin{tabular}{l|c|c|c|c|c|c|c}
	  \toprule
       \textbf{Method}                               
       & \cellcolor[HTML]{5880ab}Abs Rel 
       & \cellcolor[HTML]{5880ab}Sq Rel 
       & \cellcolor[HTML]{5880ab}RMSE  
       & \cellcolor[HTML]{5880ab}RMSE$_{log}$ 
       & \cellcolor[HTML]{e8715b}$\delta < 1.25$ 
       & \cellcolor[HTML]{e8715b}$\delta < 1.25^2$ 
       & \cellcolor[HTML]{e8715b}$\delta < 1.25^3$\\
	   \toprule
	   FisheyeDistanceNet~\cite{kumar2020fisheyedistancenet} & 0.152 & 0.768 & 2.723 & 0.210 & 0.812 & 0.954 & 0.974 \\
	   SynDistNet fixed $\alpha=1$                 & 0.148 & 0.642 & 2.615 & 0.203 & 0.824 & 0.960 & 0.978 \\
	   SynDistNet fixed $\alpha=0$                 & 0.151 & 0.638 & 2.601 & 0.205 & 0.822 & 0.962 & 0.981 \\
	   SynDistNet fixed $\alpha=2$                 & 0.154 & 0.631 & 2.532 & 0.198 & 0.832 & 0.965 & 0.981 \\
       SynDistNet adaptive $\alpha \in (0, 2)$  & \textbf{0.142} & \textbf{0.537} & \textbf{2.316} & \textbf{0.179} & \textbf{0.878} & \textbf{0.971} & \textbf{0.985} \\
	   \bottomrule
       \end{tabular}
    \end{center}
% \caption{Ablation study on different variants of our SynDistNet using the Fisheye WoodScape dataset~\cite{yogamani2019woodscape}. Distances are capped at $40\,m$. By replacing the \lone loss with several variants of the general loss function in which the loss’s shape parameters are fixed, annealed, or adaptive, we see a significant performance improvement. The fixed' entries all use the general loss imposed on wavelet coefficients, but for each entry we use a different strategy for setting the shape parameter or parameters. The input resolution is $512 \times 256$ pixels.}
\vspace{-15pt}
\caption{\textbf{Ablation study on different variants of our SynDistNet} using the Fisheye WoodScape dataset~\cite{yogamani2019woodscape}. We replace the \lone loss with several variants of the general loss function varying the parameter $\alpha$ and observe a significant performance improvement. }
\label{table:robust-ablation}
\end{table*}
% -------------------------------------------------

Table~\ref{tab:mtl} captures the primary goal of this paper, which is to develop a synergistic multi-task network for semantic segmentation and distance estimation tasks. ResNet18 encoder was used in these experiments on the Fisheye WoodScape dataset. Firstly, we formulate single-task baselines for these tasks and build an essential shared encoder multi-task learning (MTL) baseline. The MTL results are slightly better than their respective single-task benchmarks demonstrating that shared encoder features can be learned for diverse tasks wherein segmentation captures semantic features, and distance estimation captures geometric features. The proposed synergized MTL network SynDistNet reduces distance RMSE by $25\%$ and improves segmentation accuracy by $4\%$. We break down these results further using extensive ablation experiments.\par

% -------------------------------------------------
\textbf{\textit{Ablation Experiments}}
For our ablation analysis, we consider two variants of ResNet encoder heads. Distance estimation results of these variants are shown in Table~\ref{table:table_ablation}.  Significant improvements in accuracy are obtained with the replacement of \lone loss with a generic parameterized loss function. The impact of the mask is incremental in the WoodScape dataset. Still, it poses the potential to solve the infinite depth/distance issue and provides a way to improve the photometric loss. We can see with the addition of our proposed self-attention based encoder coupled with semantic-guidance decoder architecture can consistently improve the performance. Finally, with all our additions we outperform FisheyeDistanceNet~\cite{kumar2020fisheyedistancenet} for all considered metrics.\par

% -------------------------------------------------
\begin{table*}[!ht]
\captionsetup{belowskip=-12pt, font= small, singlelinecheck=false}
\renewcommand{\arraystretch}{0.87}
\centering
% Worst Scenario lets squeeze this fat guy
%\scalebox{0.9}{
{
\small
\setlength{\tabcolsep}{0.3em}
\begin{tabular}{c|lcccccccc}
\toprule
& \textbf{Method} 
& Resolution  
& \cellcolor[HTML]{5880ab}Abs Rel 
& \cellcolor[HTML]{5880ab}Sq Rel 
& \cellcolor[HTML]{5880ab}RMSE 
& \cellcolor[HTML]{5880ab}RMSE$_{log}$ 
& \cellcolor[HTML]{e8715b}$\delta<1.25$ 
& \cellcolor[HTML]{e8715b}$\delta<1.25^2$ 
& \cellcolor[HTML]{e8715b}$\delta<1.25^3$ \\
\cmidrule(lr){4-7} \cmidrule(lr){8-10}
& & & \multicolumn{4}{c}{\cellcolor[HTML]{5880ab}lower is better} & \multicolumn{3}{c}{\cellcolor[HTML]{e8715b}higher is better} \\
\midrule
  \multicolumn{10}{c}{\cellcolor[HTML]{448BE9}\textit{KITTI}}  \\
\midrule
\parbox[t]{2mm}{\multirow{10}{*}{\rotatebox[origin=c]{90}{Original~\cite{Eigen_14}}}}
& EPC++~\cite{Luo2019a}                       & 640 x 192 & 0.141 & 1.029 & 5.350 & 0.216 & 0.816 & 0.941 & 0.976 \\
& Monodepth2~\cite{Godard2019}                & 640 x 192 & 0.115 & 0.903 & 4.863 & 0.193 & 0.877 & 0.959 & 0.981 \\
& PackNet-SfM~\cite{Guizilini2020a}           & 640 x 192 & 0.111 & 0.829 & 4.788 & 0.199 & 0.864 & 0.954 & 0.980 \\
& FisheyeDistanceNet~\cite{kumar2020fisheyedistancenet} & 640 x 192 & 0.117 & 0.867 & 4.739 & 0.190 & 0.869 & 0.960 & 0.982 \\
& UnRectDepthNet~\cite{kumar2020unrectdepthnet} & 640 x 192 & \textbf{0.107} & 0.721 & 4.564 & \textbf{0.178} & 0.894 & 0.971 & \textbf{0.986} \\
& \textbf{SynDistNet}                         & 640 x 192 & 0.109 & \textbf{0.718} & \textbf{4.516} & 0.180 & \textbf{0.896} & \textbf{0.973} & \textbf{0.986} \\
\cmidrule{2-10} 
& Monodepth2~\cite{Godard2019}                & 1024 x 320 & 0.115 & 0.882 & 4.701 & 0.190 & 0.879 & 0.961 & 0.982 \\
& FisheyeDistanceNet~\cite{kumar2020fisheyedistancenet} & 1024 x 320 & 0.109 & 0.788 & 4.669 & 0.185 & 0.889 & 0.964 & 0.982 \\
& UnRectDepthNet~\cite{kumar2020unrectdepthnet} & 1024 x 320 & 0.103 & 0.705 & 4.386 & \textbf{0.164 }& 0.897 & \textbf{0.980} & 0.989 \\
& \textbf{SynDistNet}                         & 1024 x 320   & \textbf{0.102} & \textbf{0.701} & \textbf{4.347} & 0.166 & \textbf{0.901} & \textbf{0.980} & \textbf{0.990} \\
\midrule
\parbox[t]{2mm}{\multirow{8}{*}{\rotatebox[origin=c]{90}{Improved~\cite{uhrig2017sparsity}}}}
& SfMLeaner~\cite{zhou2017unsupervised}       & 416 x 128  & 0.176 & 1.532 & 6.129 & 0.244 & 0.758 & 0.921 & 0.971 \\
& Vid2Depth~\cite{mahjourian2018unsupervised} & 416 x 128  & 0.134 & 0.983 & 5.501 & 0.203 & 0.827 & 0.944 & 0.981 \\
& DDVO~\cite{Wang2018e}                       & 416 x 128  & 0.126 & 0.866 & 4.932 & 0.185 & 0.851 & 0.958 & 0.986 \\
\cmidrule{2-10}
& EPC++~\cite{Luo2019a}                     & 640 x 192  & 0.120 & 0.789 & 4.755 & 0.177 & 0.856 & 0.961 & 0.987 \\
& Monodepth2~\cite{Godard2019}              & 640 x 192  & 0.090 & 0.545 & 3.942 & 0.137 & 0.914 & 0.983 & 0.995 \\
& PackNet-SfM~\cite{Guizilini2020a}         & 640 x 192  & 0.078 & 0.420 & 3.485 & 0.121 & \textbf{0.931} & 0.986 & \textbf{0.996} \\
& UnRectDepthNet~\cite{kumar2020unrectdepthnet} & 640 x 192 & 0.081 & 0.414 & 3.412 & 0.117 & 0.926 & 0.987 & \textbf{0.996} \\
& \textbf{SynDistNet}                       & 640 x 192 & \textbf{0.076} & \textbf{0.412} & \textbf{3.406} & \textbf{0.115} & \textbf{0.931} & \textbf{0.988} & \textbf{0.996} \\
\bottomrule
\end{tabular}
}
\vspace{-5pt}
\caption{\textbf{Quantitative performance comparison of our network} with other self-supervised monocular methods for depths up to 80\,m for KITTI. \textit{Original} uses raw depth maps as proposed by \cite{Eigen_14} for evaluation, and \textit{Improved} uses annotated depth maps from \cite{uhrig2017sparsity}. At test-time, all methods excluding FisheyeDistanceNet, PackNet-SfM and \textbf{Ours}, scale the estimated depths using median ground-truth LiDAR depth.}
% \caption{\textbf{Quantitative performance comparison of our network} for depths up to 80\,m for KITTI and 40\,m on the Woodscape dataset \cite{yogamani2019woodscape}. \textit{Original} uses raw depth maps as proposed by \cite{Eigen_14} for evaluation, and \textit{Improved} uses annotated depth maps from \cite{uhrig2017sparsity}. All the approaches are self-supervised on monocular video sequences. At test-time, all monocular methods excluding FisheyeDistanceNet, PackNet-SfM and \textbf{Ours}, scale the estimated depths using median ground-truth LiDAR depth.}
\label{tab:results}
\end{table*}

% -------------------------------------------------
\begin{figure*}[htpb]
\vspace{0.15cm}
  \captionsetup{skip=-2pt, belowskip=-12pt, singlelinecheck=false, font=small}
  \centering
  \resizebox{\textwidth}{!}{
  \newcommand{\turnheightnew}{0.25\columnwidth}
\centering

\begin{tabular}{@{\hskip 0.5mm}c@{\hskip 0.5mm}c@{\hskip 0.5mm}c@{\hskip 0.5mm}c@{\hskip 0.5mm}c@{}}

{\rotatebox{90}{\hspace{4mm}Raw Input}} &
\includegraphics[height=\turnheightnew]{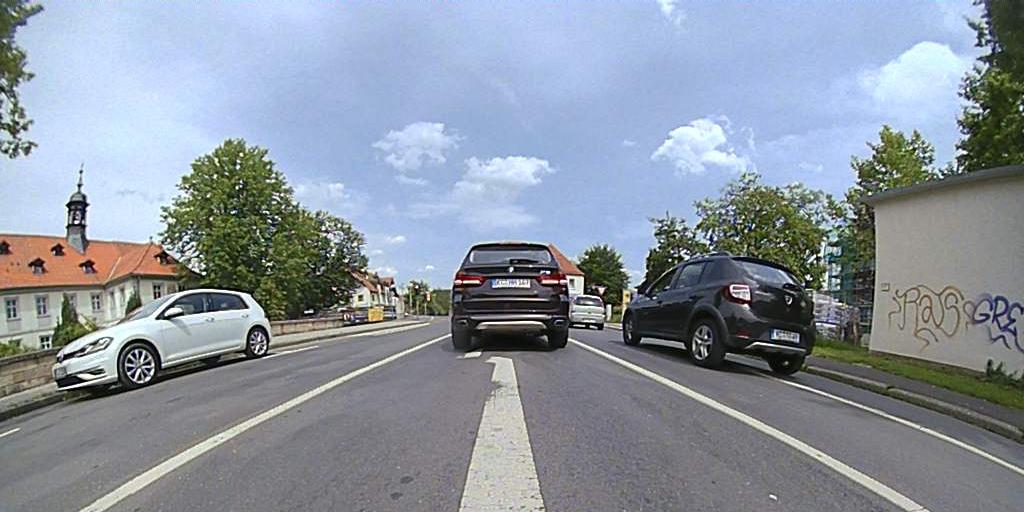} &
\includegraphics[height=\turnheightnew]{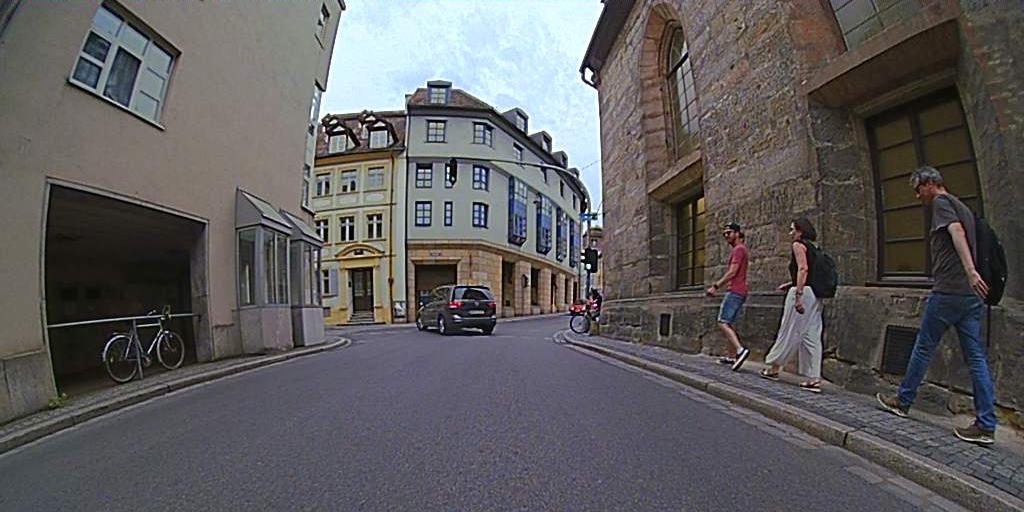} &
\includegraphics[height=\turnheightnew]{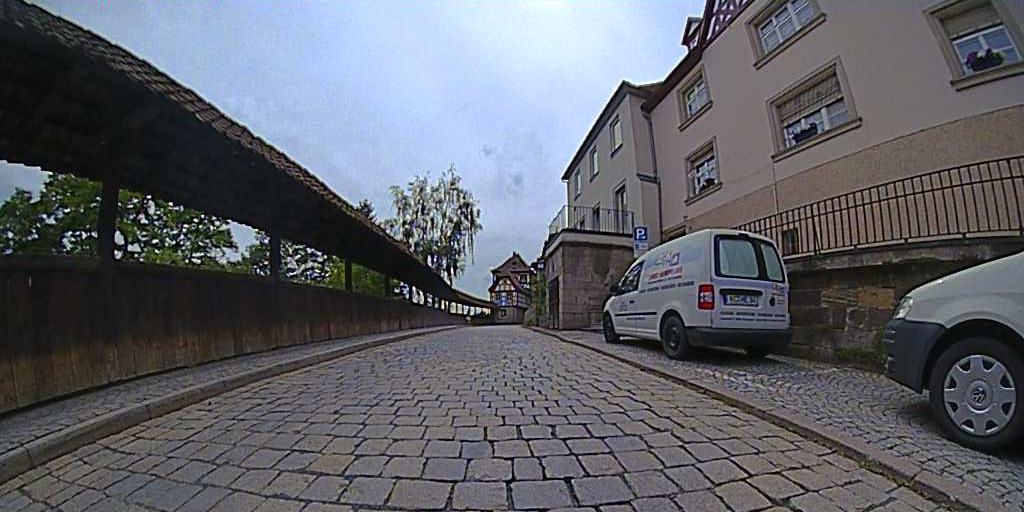} &
\includegraphics[height=\turnheightnew]{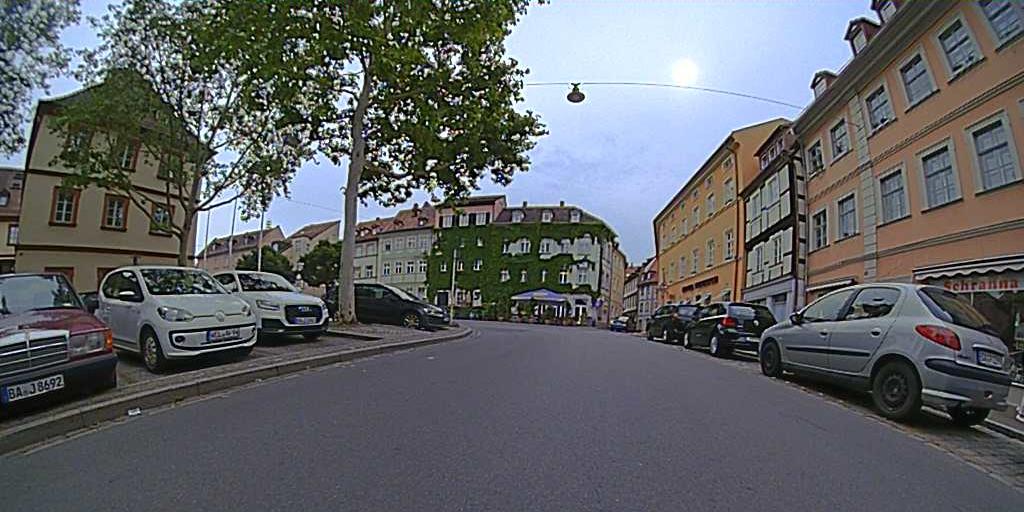}\\

{\rotatebox{90}{\hspace{4mm}Baseline}} &
\includegraphics[height=\turnheightnew]{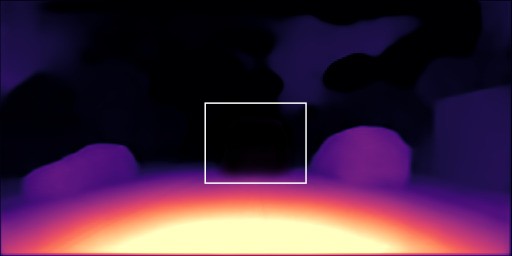} &
\includegraphics[height=\turnheightnew]{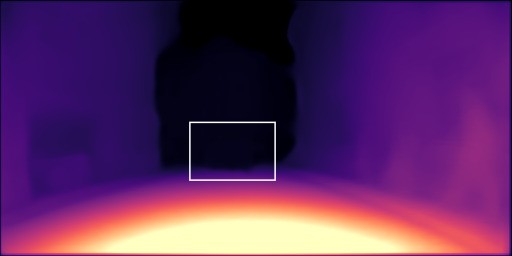} &
\includegraphics[height=\turnheightnew]{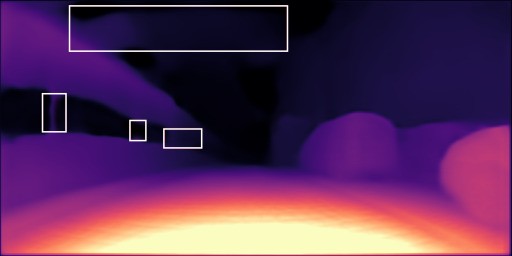} &
\includegraphics[height=\turnheightnew]{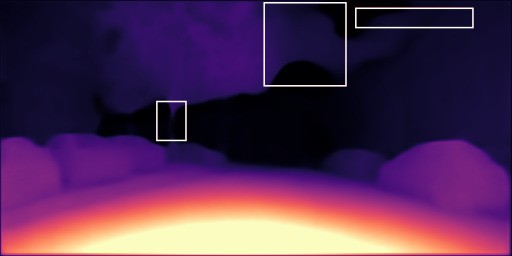}\\

{\rotatebox{90}{\hspace{3mm}SynDistNet}} &
\includegraphics[height=\turnheightnew]{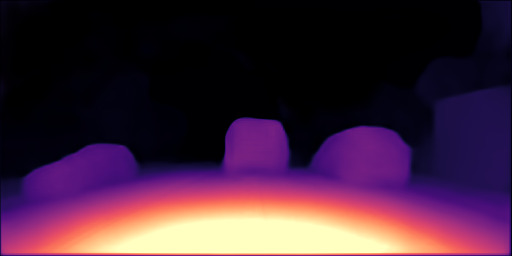} &
\includegraphics[height=\turnheightnew]{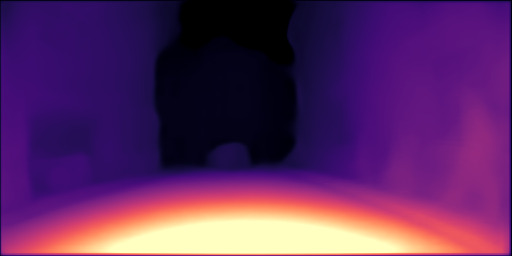} &
\includegraphics[height=\turnheightnew]{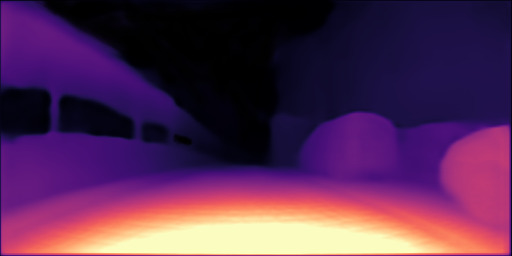} &
\includegraphics[height=\turnheightnew]{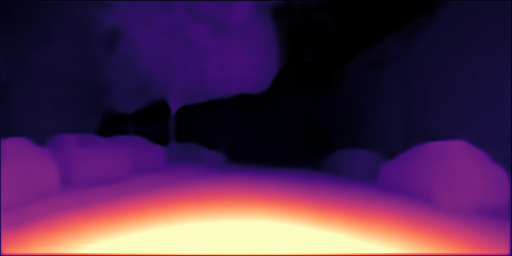} \\

\end{tabular}}
  \caption{\textbf{Qualitative result comparison on the Fisheye WoodScape dataset} between the baseline model without our contributions and the proposed SynDistNet. Our SynDistNet can recover the distance of dynamic objects (left images) which eventually solves the infinite distance issue. In the 3rd and 4th columns, we can see that semantic guidance helps us to recover the thin structure and resolve the distance of homogeneous areas outputting sharp distance maps on raw fisheye images.}
  \label{fig:fisheye_qual}
\end{figure*}

% -------------------------------------------------
\textbf{\textit{Robust loss function strategy}}
%The \lone loss is replaced with different variants of the robust general loss~\cite{barron2019general}, and we showcase that the usage of adaptive or annealed variants of the loss can significantly improve the performance.
We showcase that adaptive or annealed variants of the robust loss can significantly improve the performance.
%The shape parameter $\alpha$ is varied, keeping the scale fixed with a general distribution than a fixed Laplacian distribution. 
% Instead of RGB representation, following~\cite{barron2019general}, YUV wavelet representations are used to model the images, and the loss is applied to a YUV wavelet decomposition.
%The multi-scale training as the reconstruction loss in FisheyeDistanceNet~\cite{kumar2019fisheyedistancenet} is dropped, which induces the sum of the means of the losses imposed at each scale in a $D$-level pyramid of side prediction since~\cite{barron2019general} is a $D$ level normalized wavelet decomposition. 
Compared to~\cite{barron2019general} we retained the edge smoothness loss from FisheyeDistanceNet~\cite{kumar2020fisheyedistancenet} as it yielded better results. The fixed scale assumption is matched by setting the loss's scale $c$ fixed to $0.01$, which also roughly matches the shape of its \lone loss.
%The loss is multiplied by $c$ to avoid exploding gradients, which bounds the gradient magnitudes by residual magnitudes.
For the fixed scale models in Table~\ref{table:robust-ablation}, we used a constant value for $\alpha$.
%We observe that there is an improvement in the performance, and there is no single value of $\alpha$, which is optimal. 
In the adaptive $\alpha \in (0, 2)$ variant, $\alpha$ is made a free parameter and is allowed to be optimized along with the network weights during training. The adaptive plan of action outperforms the fixed strategies, which showcases the importance of allowing the model to regulate the robustness of its loss during training adaptively.\par 
% A comparison of the adaptive model's performance with the fixed models indicates that no single set of $\alpha$ is optimal for all wavelet coefficients.\par
% -------------------------------------------------
\textbf{\textit{KITTI Evaluation}} As there is little work on fisheye distance estimation, we evaluate our method on extensively used KITTI dataset using the metrics proposed by Eigen et al.~\cite{Eigen_14} to facilitate comparison. The quantitative results are shown in the Table~\ref{tab:results} illustrate that the improved scale-aware self-supervised approach outperforms all the state-of-the-art monocular approaches. More specifically, we improve the baseline \emph{FisheyeDistanceNet} with the usage of a general and adaptive loss function~\cite{barron2019general} which is showcased in Table~\ref{table:robust-ablation} and better architecture. We could not leverage the Cityscapes dataset into our training regime to benchmark our scale-aware framework due to the absence of odometry data. Compared to PackNet-SfM~\cite{Guizilini2020a}, which presumably uses a superior architecture than our ResNet18, where they estimate scale-aware depths with their velocity supervision loss using the ground truth poses for supervision. We only rely on speed and time data captured from the vehicle odometry, which is easier to obtain. Our approach can be easily transferred to the domain of aerial robotics as well. We could achieve higher accuracy than PackNet, which can be seen in Table~\ref{tab:results}.
% -------------------------------------------------
\vspace{-0.5cm}
% -------------------------------------------------
\section{Conclusion}
\label{sec:conclusion}
\vspace{-4pt}
Geometry and appearance are two crucial cues of scene understanding, \eg, in automotive scenes. In this work, we develop a multi-task learning model to estimate metric distance and semantic segmentation in a synergized manner. Specifically, we leverage the semantic segmentation of potentially moving objects to remove wrongful projected objects inside the view synthesis step. We also propose a novel architecture to semantically guide the distance estimation that is trainable in a one-stage fashion and introduce the application of a robust loss function. Our primary focus is to develop our proposed model for less explored fisheye cameras based on the WoodScape dataset. We demonstrate the effect of each proposed contribution individually and obtain state-of-the-art results on both WoodScape and KITTI datasets for self-supervised distance estimation.
% -------------------------------------------------
\vspace{-3em}
\bibliographystyle{ieee_fullname}
{\small
\bibliography{bib/egbib}
}

\clearpage
\setcounter{section}{0}
\begin{center}
    {\Large \textsc{Supplementary Material}}
\end{center}

% -------------------------------------------------
\section{Additional Method Details}

\textbf{\textit{Edge-Aware Distance Smoothness Loss}}:
In order to regularize distance and avoid divergent values in occluded or texture-less low-image gradient areas, we add a geometric smoothing loss. We adopt the edge-aware term similar to~\cite{Godard2019}. The regularization term is imposed on the inverse distance map. The loss is weighted for each of the image pyramid levels and is decayed by a factor of 2 on each downsampling.
\begin{equation}
    \mathcal{L}_{s}(\hat{D}_t) = | \partial_u \hat{D}^*_t | e^{-|\partial_u I_t|} + | \partial_v \hat{D}^*_t | e^{-|\partial_v I_t|}
\end{equation}
To discourage shrinking of distance estimates~\cite{Wang2018e}, mean-normalized inverse distance of $D_t$ is considered, i.e. $\hat{D}^*_t = \hat{D}^{-1}_t / \overline{D}_t$, where $\overline{D}_t$ denotes the mean of $\hat{D}^{-1}_t := 1 /\hat{D}_t$.\par
% -------------------------------------------------
\textbf{\textit{Cross-Sequence Distance Consistency Loss}}:
Following FisheyeDistanceNet~\cite{kumar2020fisheyedistancenet}, we enforce the cross-sequence distance consistency loss (CSDCL) for the training sequence~$S$:
\begin{align} 
    \label{equ:dclf}
    \mathcal{L}_{dc} = \sum_{t=1}^{N-1} \sum_{t'=t+1}^{N} \bigg( &\sum_{p_t}
    \left| D_{t \to t^\prime}\left(p_t \right) - \hat D_{t \to t^\prime}\left(p_t \right) \right| \nonumber \\
    + &\sum_{p_{t'}} \left| D_{t' \to t}\left(p_{t'} \right) - \hat D_{t' \to t}\left(p_{t'} \right) \right| \bigg)
\end{align}
Eq. \ref{equ:dclf} contains one term for which pixels and point clouds are warped forwards in time (from $t$ to $t'$) and one term for which they are warped backwards in time (from $t'$ to $t$), where $\hat D_{t'}$ and $\hat D_{t}$ are the estimates of the images $I_{t'}$ and $I_t$ respectively for each pixel ${{p}_{t}}\in {{I}_{t}}$.\par
% -------------------------------------------------
\begin{figure}[!ht]
  \captionsetup{belowskip=-8pt, singlelinecheck=false, font=small}
  \centering
  \resizebox{\columnwidth}{!}{
  \newcommand{\turnheightnew}{0.25\columnwidth}
\centering

\begin{tabular}{@{\hskip 0.5mm}c@{\hskip 0.5mm}c@{\hskip 0.5mm}c@{\hskip 0.5mm}c@{\hskip 0.5mm}c@{}}

{\rotatebox{90}{\hspace{4mm}Raw Input}} &
\includegraphics[height=\turnheightnew]{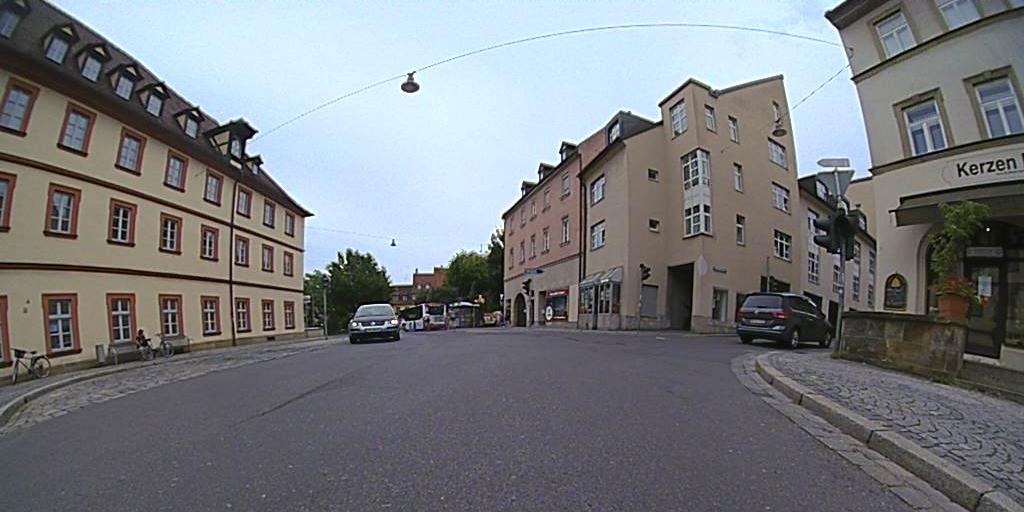} &
\includegraphics[height=\turnheightnew]{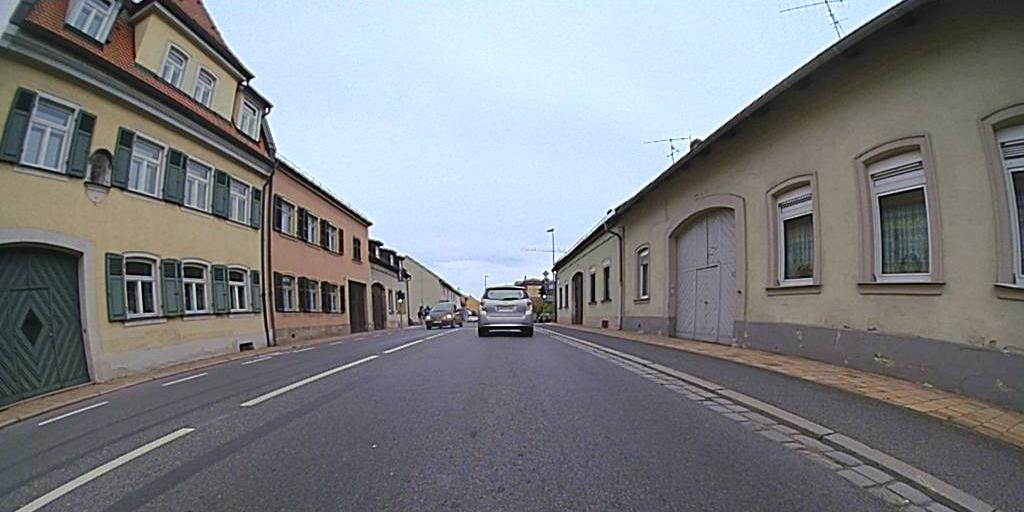}\\

{\rotatebox{90}{\hspace{4mm}Baseline}} &
\includegraphics[height=\turnheightnew]{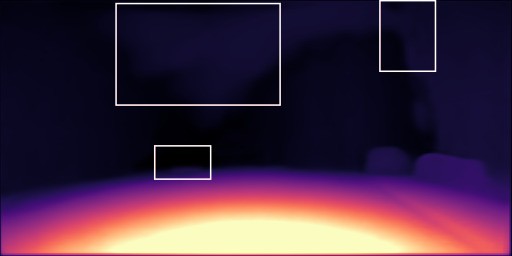} &
\includegraphics[height=\turnheightnew]{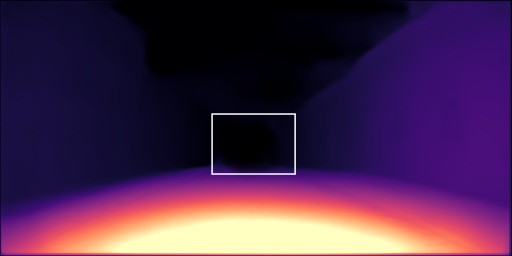}\\

{\rotatebox{90}{\hspace{3mm}SynDistNet}} &
\includegraphics[height=\turnheightnew]{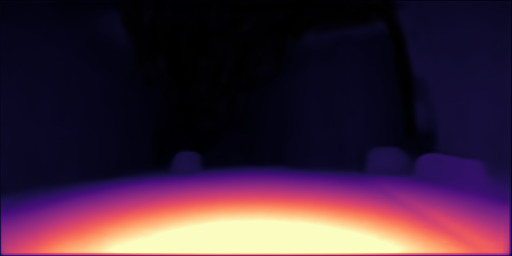} &
\includegraphics[height=\turnheightnew]{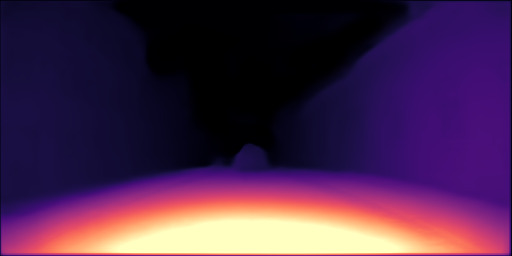} \\

{\rotatebox{90}{\hspace{3mm}SynDistNet}} &
\includegraphics[height=\turnheightnew]{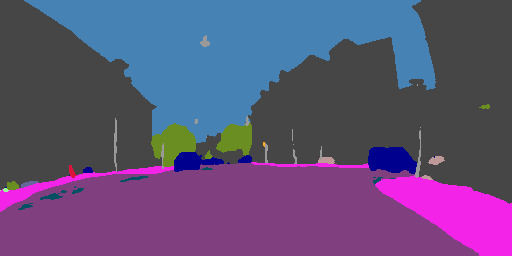} &
\includegraphics[height=\turnheightnew]{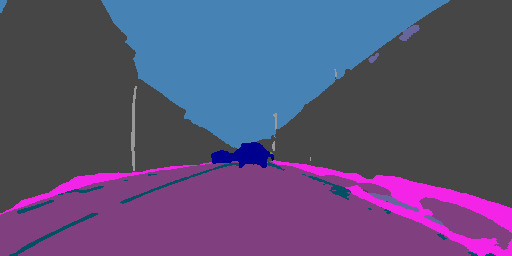} \\

\end{tabular}
  }
  \caption{\textbf{Qualitative result comparison on the Fisheye WoodScape dataset} between the baseline model without our contributions and the proposed SynDistNet. Our SynDistNet can recover the distance of dynamic objects (left images) which eventually solves the infinite distance issue. In the 3rd and 4th columns, we can see that semantic guidance helps us to obtain curbs and resolve the distance of homogeneous areas outputting sharp distance maps on raw fisheye images. The final row indicates the semantic segmentation predictions.}
  \label{fig:WoodscapeMTLResults}
\end{figure}
% -------------------------------------------------
\textbf{\textit{Additional Considerations}}:
In all the previous works~\cite{zhou2017unsupervised, Godard2019, Guizilini2020a}, networks are trained to recover inverse depth $g_d: p \mapsto  g^{-1}_D(I_t(p))$. A limitation of these approaches is that both depth or distance and pose are estimated up to an unknown scale factor. We incorporate the scale recovery technique from FisheyeDistanceNet~\cite{kumar2020fisheyedistancenet} and obtain scale-aware depth and distance directly for pinhole and fisheye images. We also incorporate the clipping of the photometric loss values, which improves the optimization process and provides a way to strengthen the photometric error. Additionally, we include the backward sequence training regime, which helps to resolve the unknown distance estimates in the image border.\par
% -------------------------------------------------

\section{Implementation Details}

% -------------------------------------------------
\begin{figure*}[ht]
    \captionsetup{belowskip=-8pt, font= small, singlelinecheck=false}
  \resizebox{\textwidth}{!}{
  \newcommand{\turnheightnew}{0.25\columnwidth}
\centering

\begin{tabular}{c@{\hskip 0.5mm}c@{\hskip 0.5mm}c@{\hskip 0.5mm}c@{\hskip 0.5mm}c@{}}

\Large Raw Input &
\Large Zhou \cite{zhou2017unsupervised} &
\Large Godard \cite{Godard2019} &
\Large SynDistNet\\

\includegraphics[height=\turnheightnew]{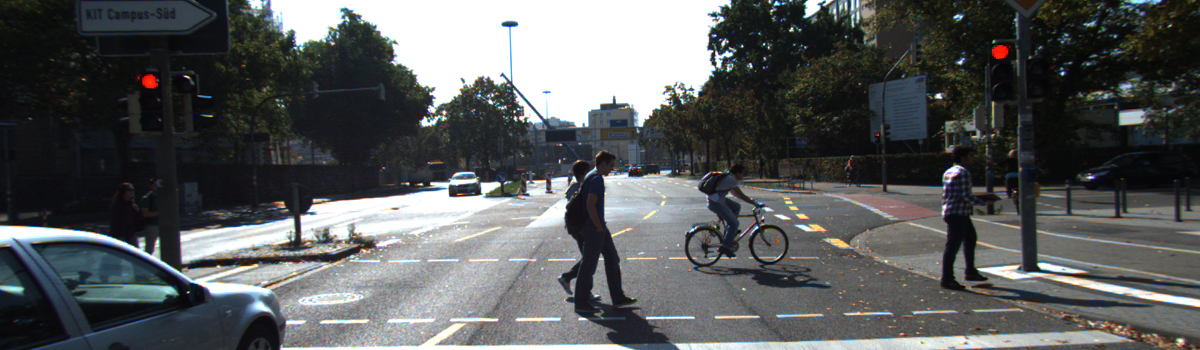}\; &
\includegraphics[height=\turnheightnew]{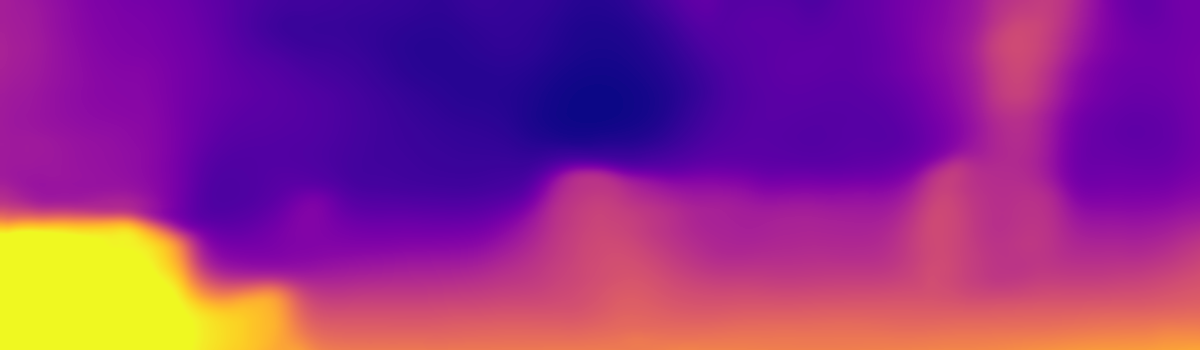}\; &
\includegraphics[height=\turnheightnew]{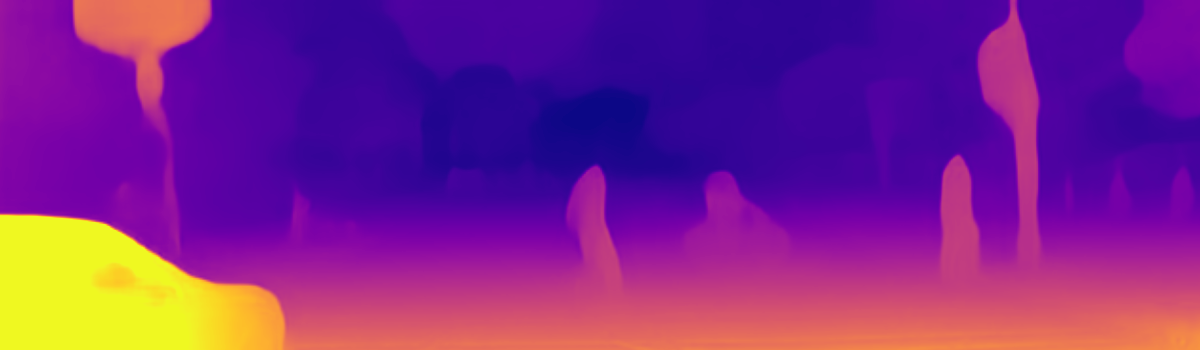}\; &
\includegraphics[height=\turnheightnew]{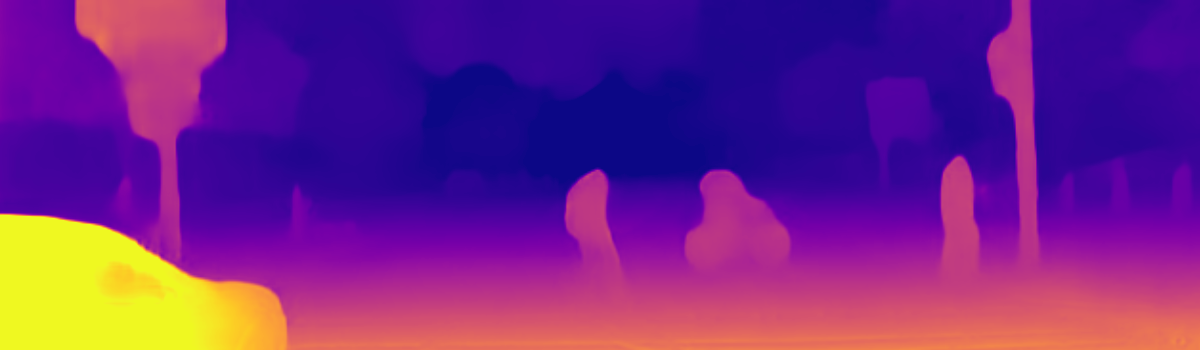}\\

\includegraphics[height=\turnheightnew]{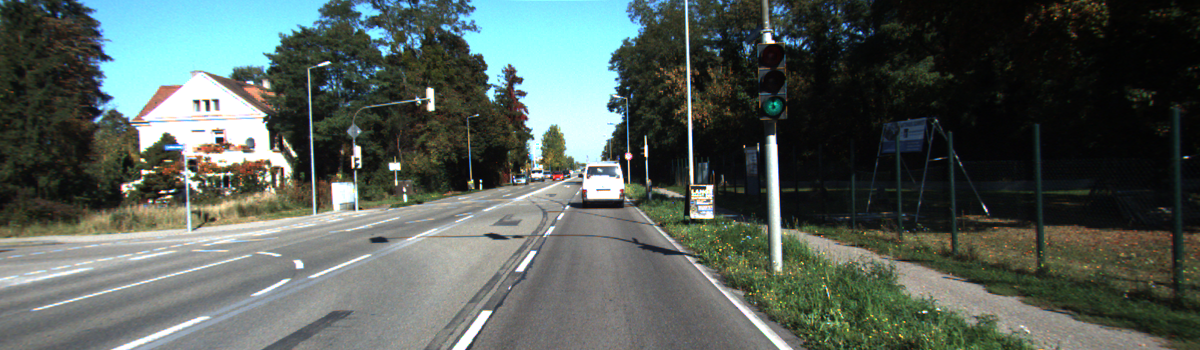}\; &
\includegraphics[height=\turnheightnew]{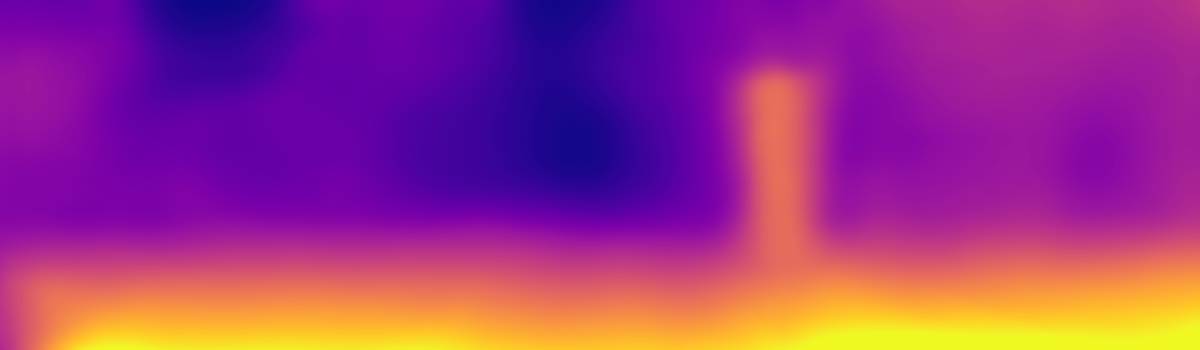}\; &
\includegraphics[height=\turnheightnew]{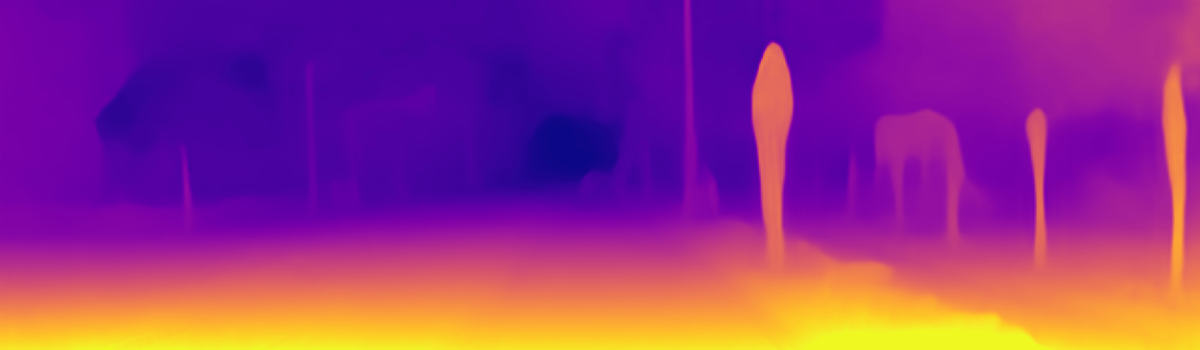}\; &
\includegraphics[height=\turnheightnew]{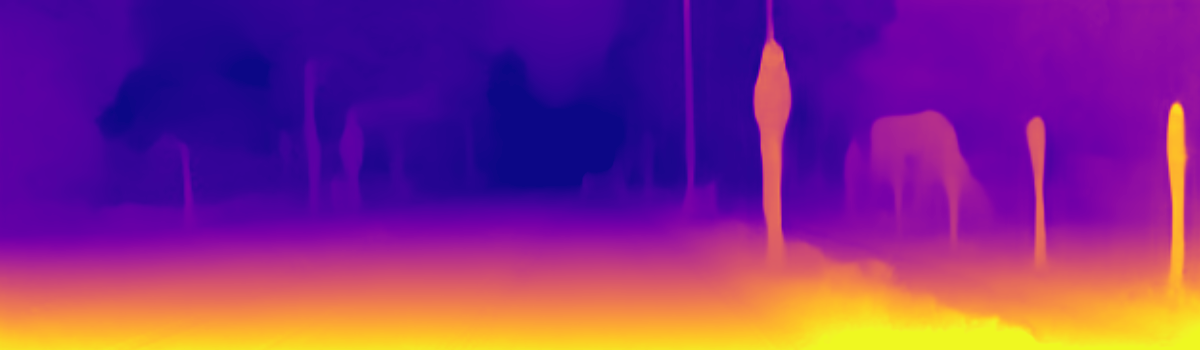}\\

\includegraphics[height=\turnheightnew]{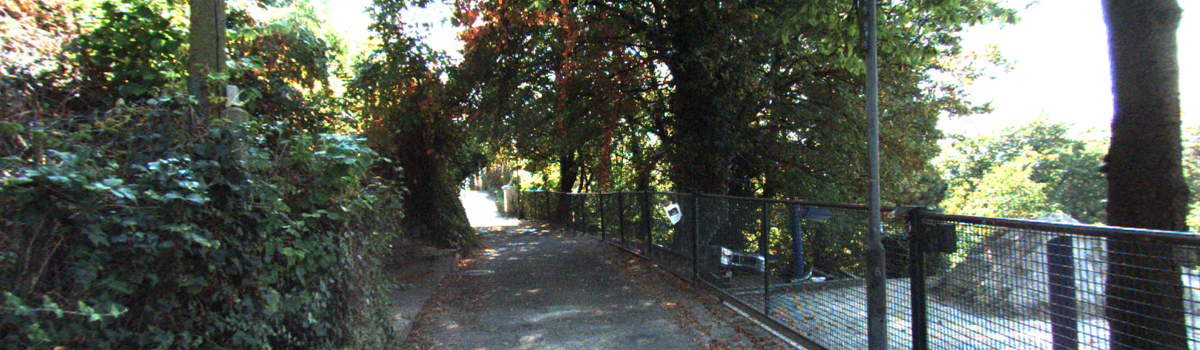}\; &
\includegraphics[height=\turnheightnew]{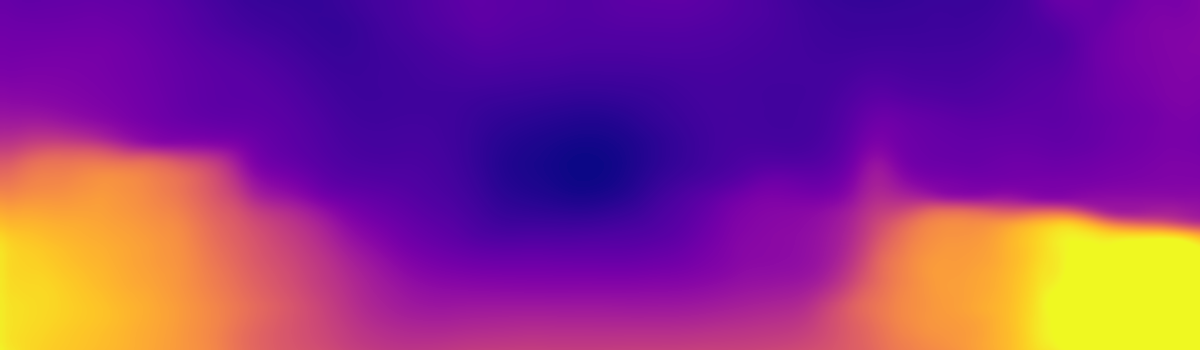}\; &
\includegraphics[height=\turnheightnew]{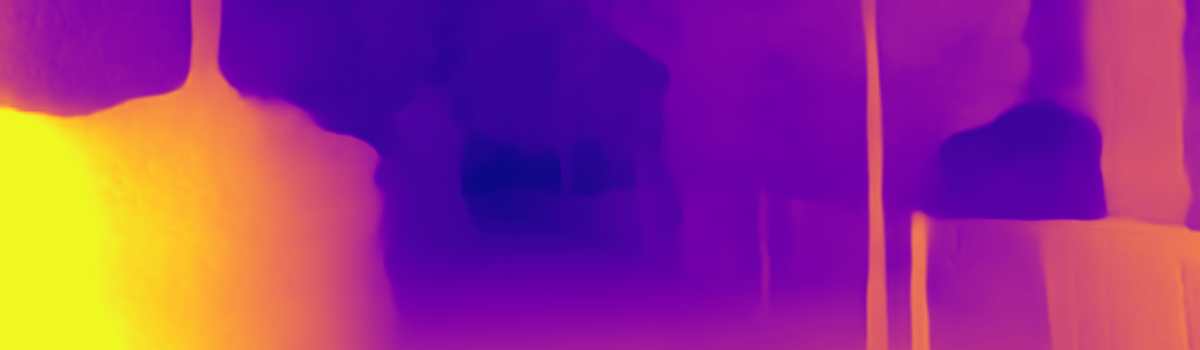}\; &
\includegraphics[height=\turnheightnew]{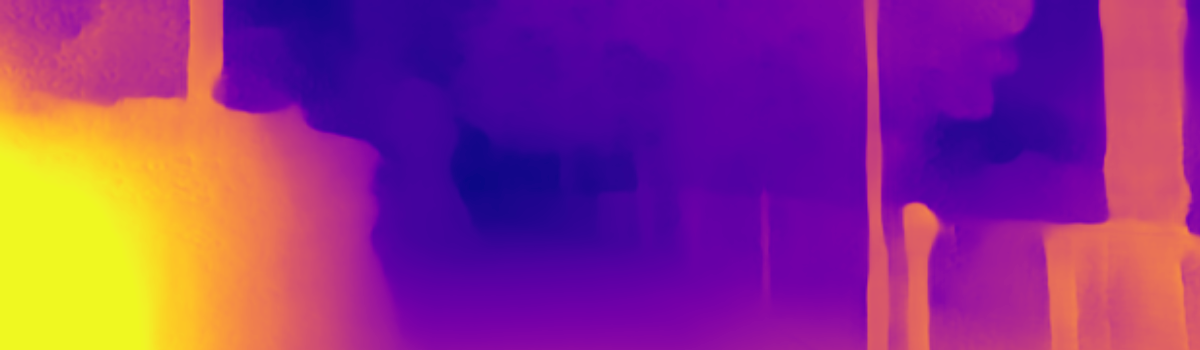}\\

 \includegraphics[height=\turnheightnew]{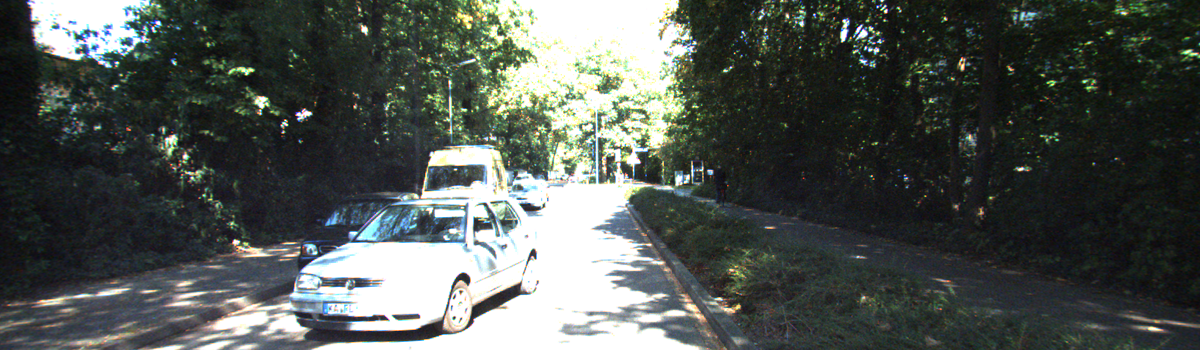}\; &
\includegraphics[height=\turnheightnew]{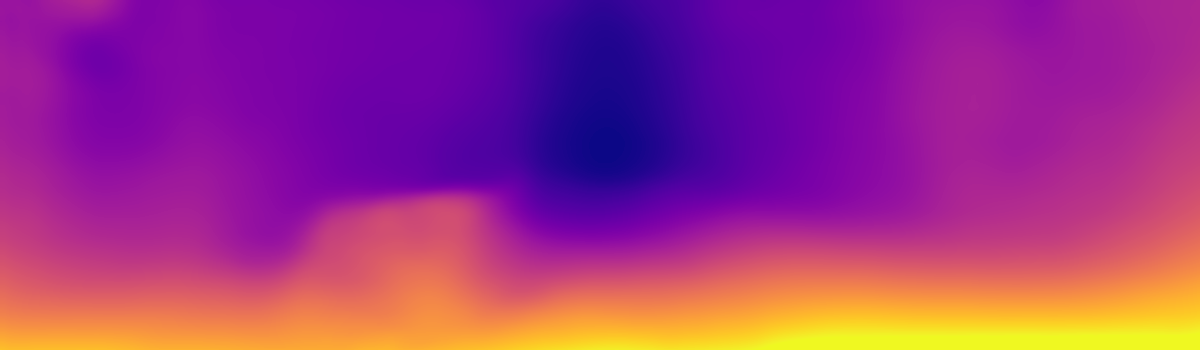}\; &
\includegraphics[height=\turnheightnew]{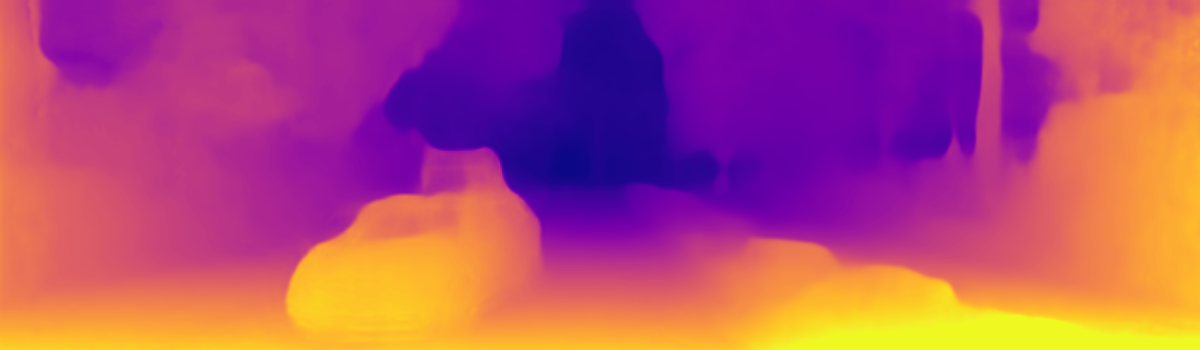}\; &
\includegraphics[height=\turnheightnew]{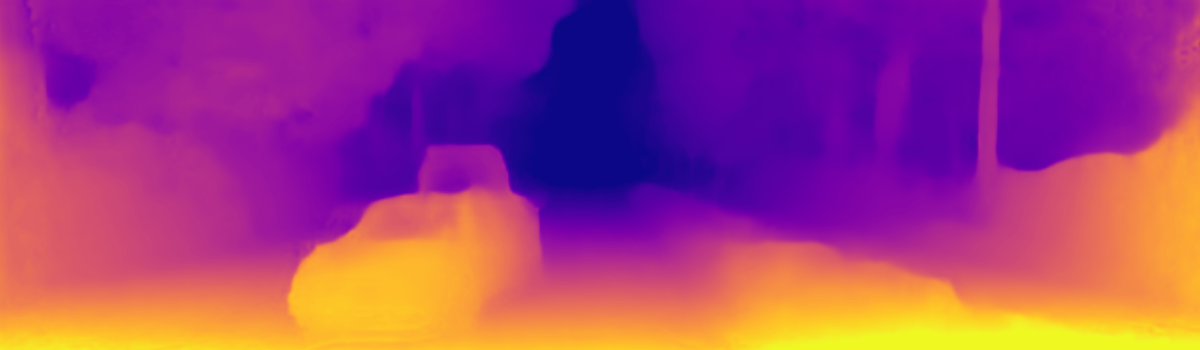}\\

\includegraphics[height=\turnheightnew]{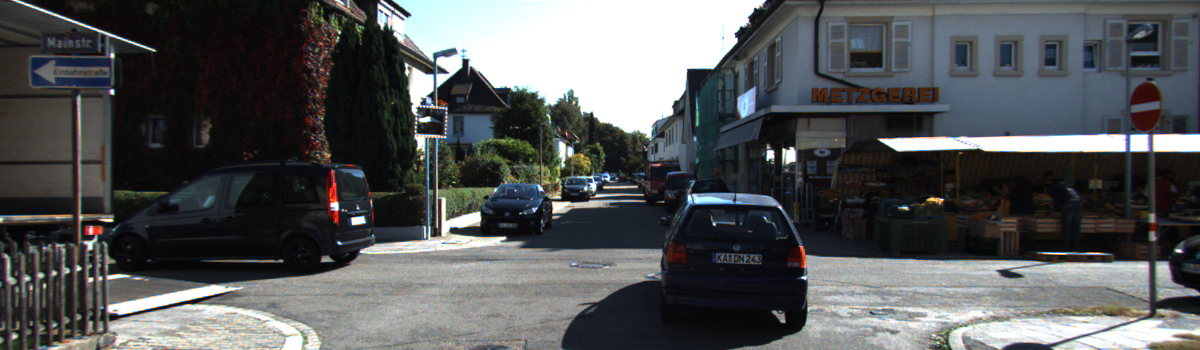}\; &
\includegraphics[height=\turnheightnew]{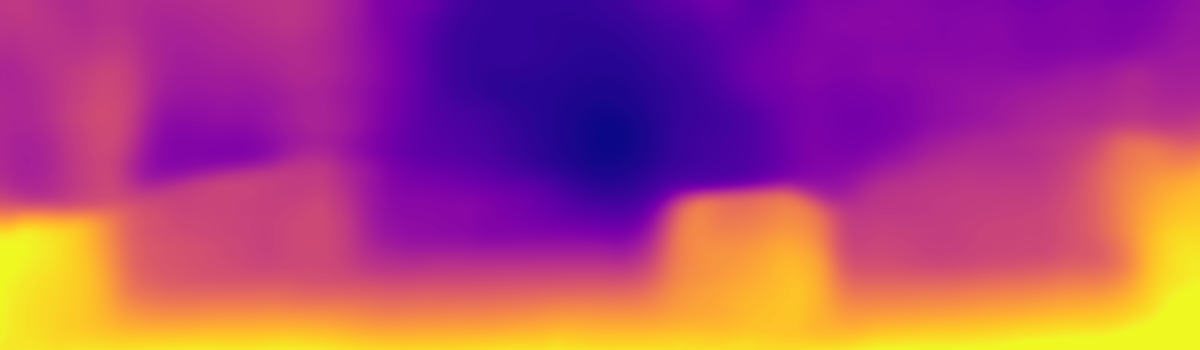}\; &
\includegraphics[height=\turnheightnew]{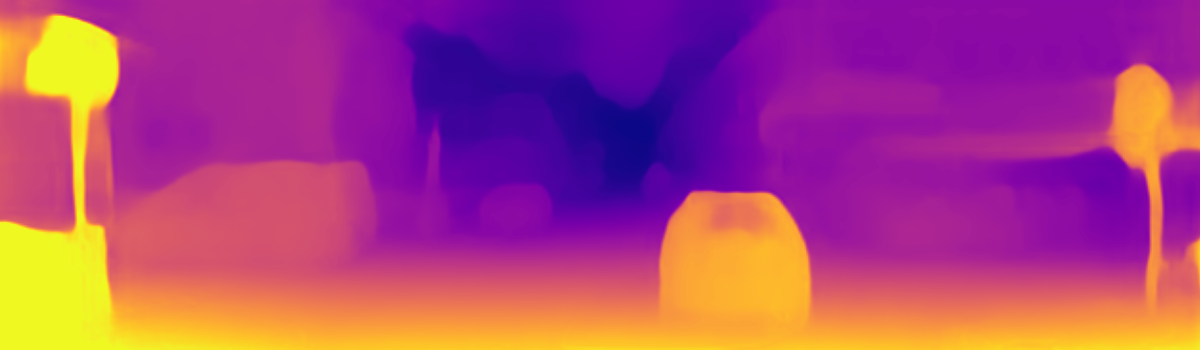}\; &
\includegraphics[height=\turnheightnew]{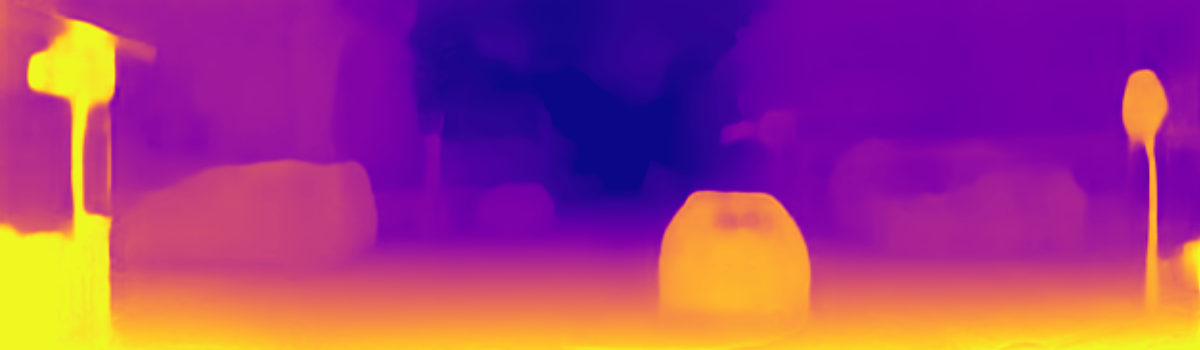}\\

\end{tabular}
  }
  \caption{\textbf{Qualitative results on the KITTI dataset.}
  Our SynDistNet produces sharp depth maps on raw pinhole camera images and can recover the distance of dynamic objects.}
  \label{fig:KITTIDepthComparison}
\end{figure*}
% -------------------------------------------------

% -------------------------------------------------
\begin{figure}[!t]
  \captionsetup{belowskip=-8pt, singlelinecheck=false, font=small}
  \centering
  \resizebox{\columnwidth}{!}{
  \newcommand{\turnheightnew}{0.25\columnwidth}
\centering

\begin{tabular}{@{\hskip 0.5mm}c@{\hskip 0.5mm}c@{\hskip 0.5mm}c@{\hskip 0.5mm}c@{\hskip 0.5mm}c@{}}

{\rotatebox{90}{\hspace{4mm}Raw Input}} &
\includegraphics[height=\turnheightnew]{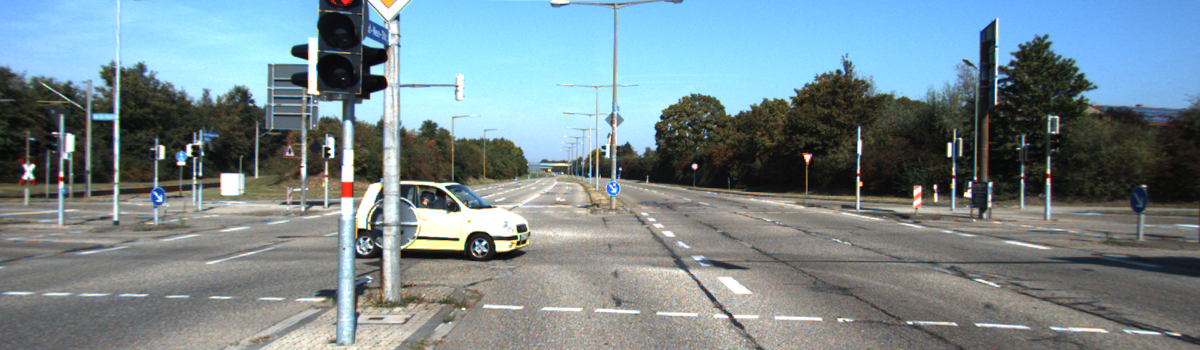} &
\includegraphics[height=\turnheightnew]{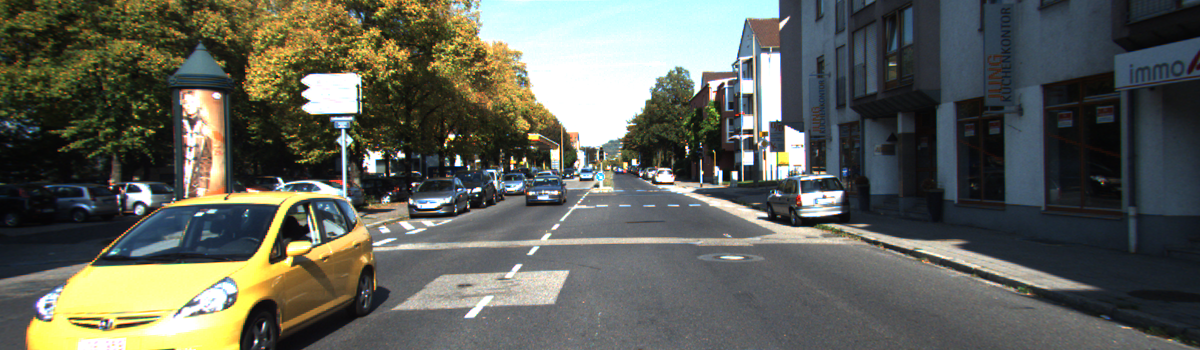} \\

{\rotatebox{90}{\hspace{4mm}SynDistNet}} &
\includegraphics[height=\turnheightnew]{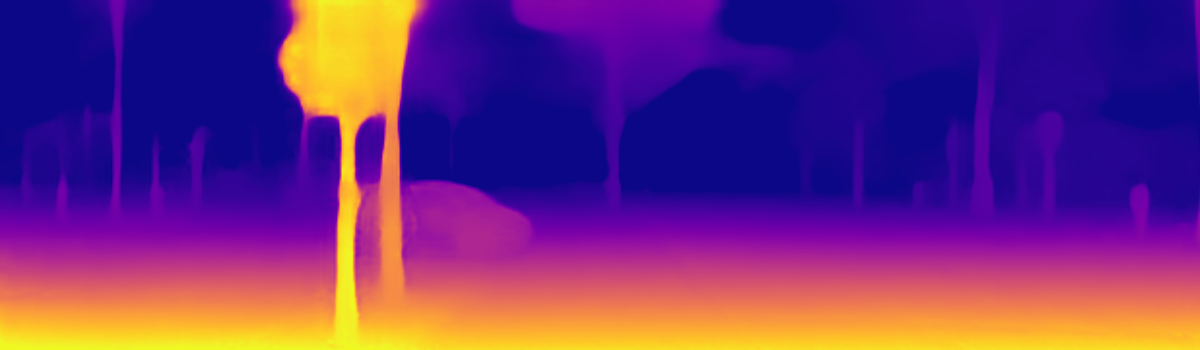} &
\includegraphics[height=\turnheightnew]{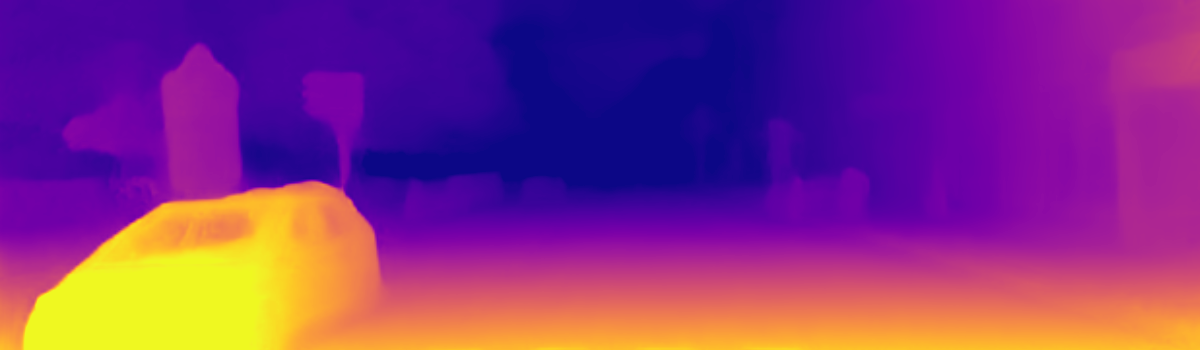} \\

{\rotatebox{90}{\hspace{3mm}SynDistNet}} &
\includegraphics[height=\turnheightnew]{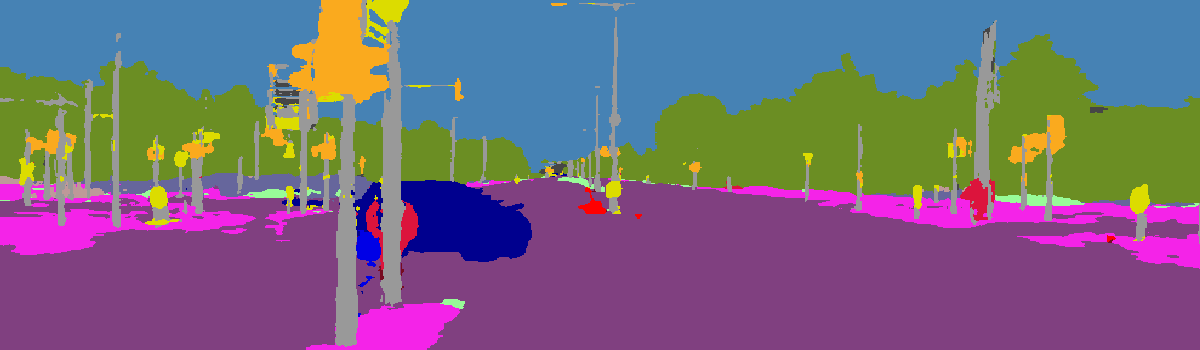} &
\includegraphics[height=\turnheightnew]{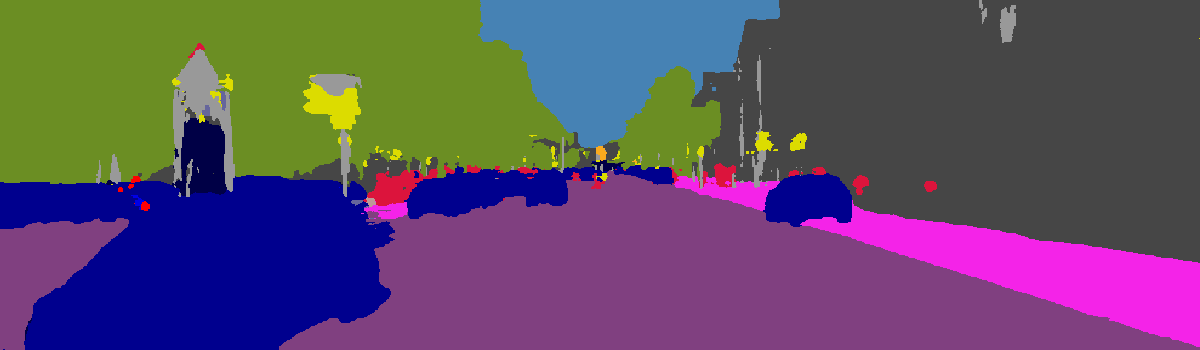} \\

{\rotatebox{90}{\hspace{4mm}Raw Input}} &
\includegraphics[height=\turnheightnew]{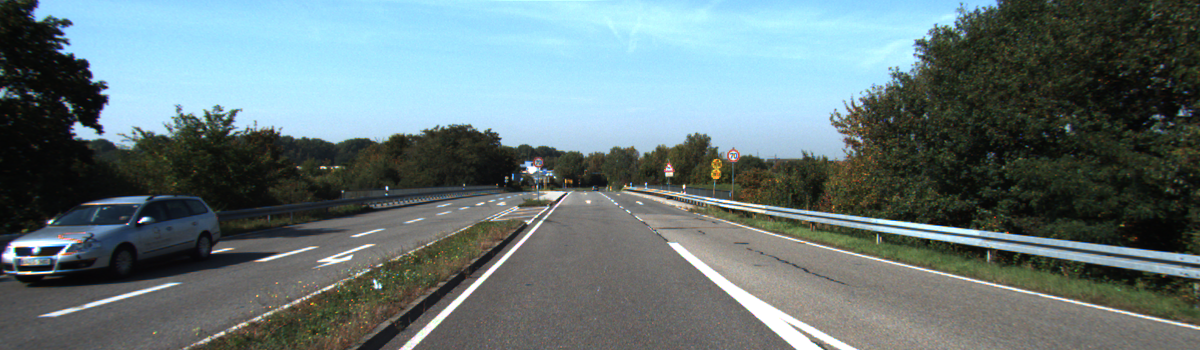} &
\includegraphics[height=\turnheightnew]{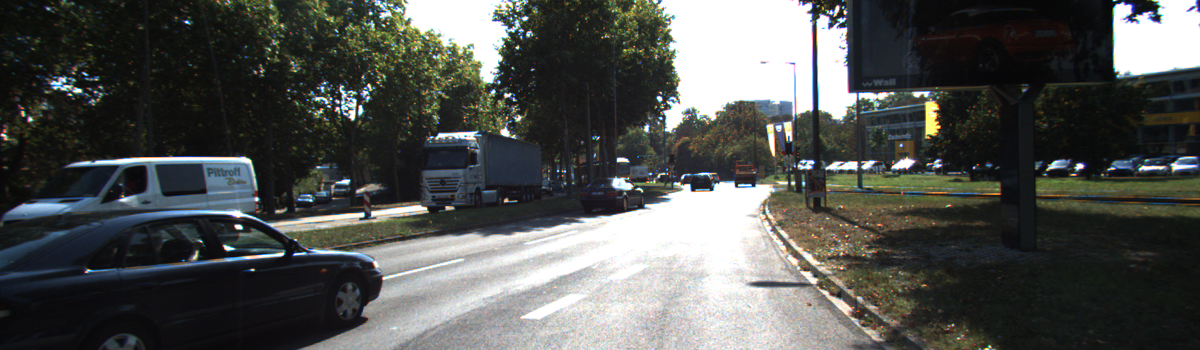} \\

{\rotatebox{90}{\hspace{4mm}SynDistNet}} &
\includegraphics[height=\turnheightnew]{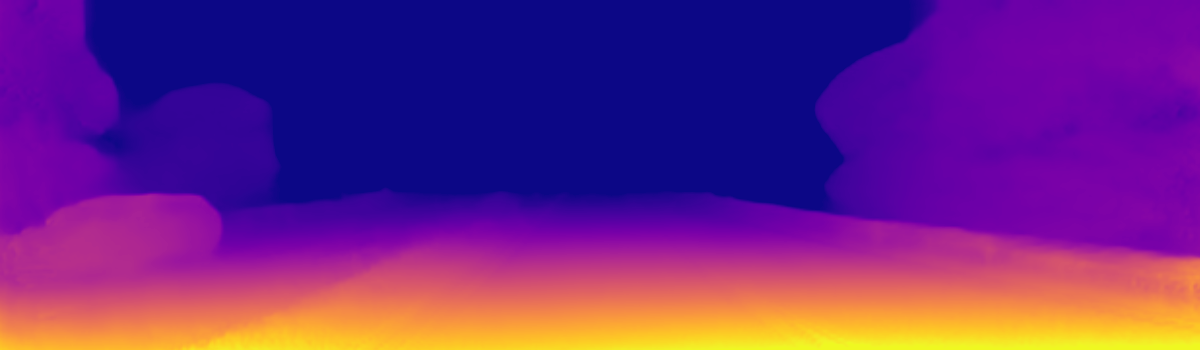} &
\includegraphics[height=\turnheightnew]{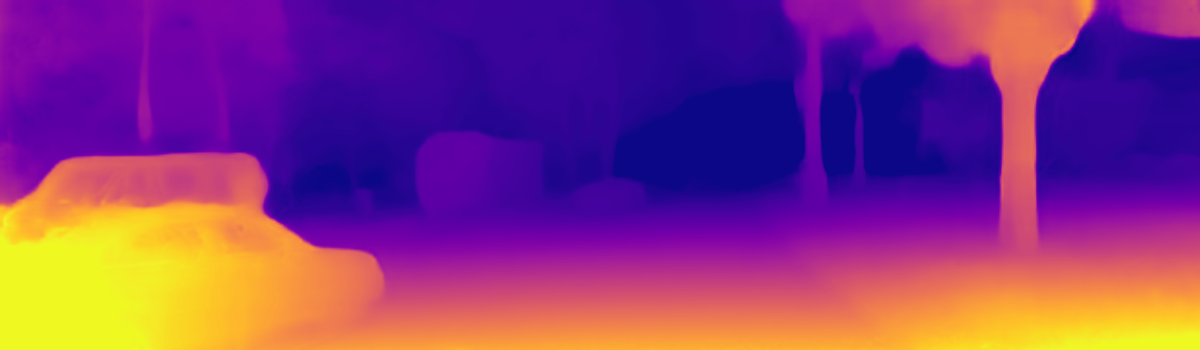} \\

{\rotatebox{90}{\hspace{3mm}SynDistNet}} &
\includegraphics[height=\turnheightnew]{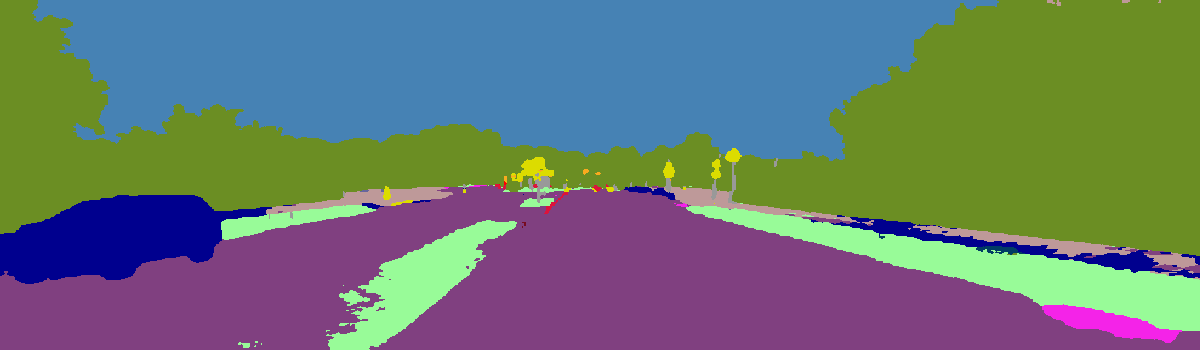} &
\includegraphics[height=\turnheightnew]{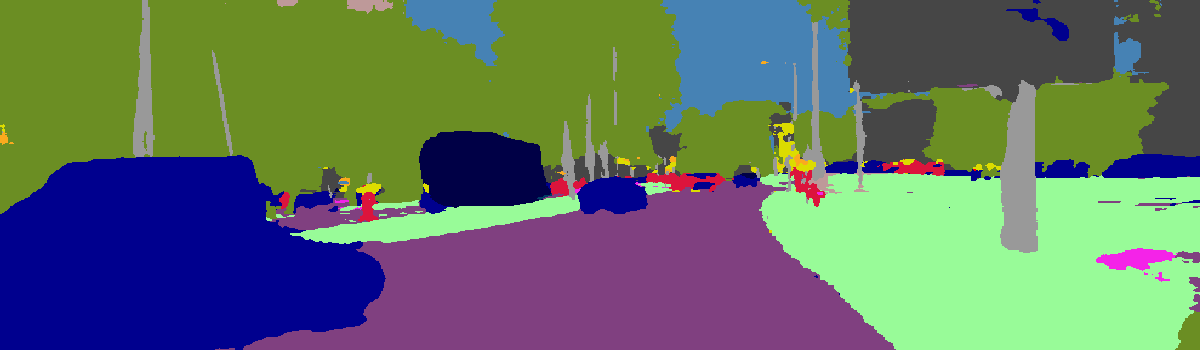} \\

\end{tabular}
  }
  \caption{\textbf{Qualitative results on the KITTI dataset.}
  We showcase depth estimation as well as semantic segmentation outputs on the KITTI dataset using our SynDistNet.}
  \label{fig:KITTIMTLResults}
\end{figure}
% -------------------------------------------------

The distance estimation network is mainly based on FisheyeDistanceNet~\cite{kumar2020fisheyedistancenet}, an \textit{encoder-decoder} network with skip connections. After testing different variants of ResNet family, we chose ResNet18~\cite{he2016deep} as the encoder as it provides a high-quality distance prediction, and improvements in higher complexity encoders were incremental. It would also aid in obtaining real-time performance on low-power embedded systems. We also incorporate self-attention layers in the encoder and drop the deformable convolutions used in the baseline model. We could leverage the usage of a more robust loss function over \lone to reduce training times on ResNet18 by performing a single-scale image depth prediction than the multi-scale in~\cite{kumar2020fisheyedistancenet}. The semantic segmentation is trained in a supervised fashion with Cross-Entropy loss and is jointly optimised along with the distance estimation. We use Pytorch~\cite{paszke2017automatic} and employ Ranger (RAdam \cite{liu2019variance} + LookAhead~\cite{zhang2019lookahead}) optimizer to minimize the training objective function than the previously employed Adam~\cite{kingma2014adam}. RAdam leverages a dynamic rectifier to adjust Adam's adaptive momentum based on the variance and effectively provides an automated warm-up custom-tailored to the current dataset to ensure a solid start to training. LookAhead "lessens the need for extensive hyperparameter tuning" while achieving "faster convergence across different deep learning tasks with minimal computational overhead." Hence, both provide breakthroughs in different aspects of deep learning optimization, and the combination is highly synergistic, possibly providing the best of both improvements for the results.\par

We train the model for 17 epochs, with a batch size of 20 on 24GB Titan RTX with an initial learning rate of ${{10}^{-4}}$ for the first 12 epochs, then drop to ${{10}^{-5}}$ for the last 5 epochs. A significant decrease in training time of 8 epochs over the previous training of the model for 25 epochs in FisheyeDistanceNet~\cite{kumar2020fisheyedistancenet}. The sigmoid output $\sigma$ from the distance decoder is converted to distance with  $D = {a \cdot \sigma + b}$. For the pinhole model, depth $D = 1 / ({a \cdot \sigma + b})$, where $a$ and $b$ are chosen to constrain $D$ between $0.1$ and $100$ units. The original input resolution of the fisheye image is $1280\times800$ pixels; we crop it to $1024\times512$ to remove the vehicle's bumper, shadow, and other artifacts of the vehicle. Finally, the cropped image is downscaled to $512\times256$ before feeding to the network. For the pinhole model on KITTI, we use $640\times192$ pixels as the network input.\par

\section{Qualitative Results}

Figure~\ref{fig:WoodscapeMTLResults} and Figure~\ref{fig:KITTIMTLResults} provides qualitative results of SynDistNet on WoodScape and KITTI test dataset for segmentation and depth estimation tasks respectively. Figure~\ref{fig:KITTIDepthComparison} illustrates the qualitative comparison of depth estimation with the recent state of the art methods.
%Additional qualitative results are provided in the two attached video files. \textit{SynDistNet\_MTL.mp4} demonstrates the performance of segmentation and distance estimation tasks on a test video captured in an unseen city in Germany. \textit{SynDistNet\_Distance.mp4} demonstrates the generalization of the distance on a fisheye camera mounted on the side of the car.

% -------------------------------------------------
\end{document}